\documentclass[Afour,sageh,times]{sagej}

\usepackage[colorlinks=true,linkcolor=black,citecolor=blue,urlcolor=blue]{hyperref}  

\usepackage{flushend}  

\newcommand{\SI}[2]{#1\,#2}
\newcommand{\percent}{\%}
\newcommand{\milli}{m}
\newcommand{\centi}{c}
\newcommand{\kilo}{k}
\newcommand{\meter}{m}
\newcommand{\second}{s}
\newcommand{\hertz}{Hz}
\newcommand{\pascal}{Pa}
\newcommand{\decibel}{dB}
\newcommand{\celsius}{$^\circ$C}

\newcommand{\hfilll}{\hspace*{\fill}}

\usepackage{fancyhdr}
\fancypagestyle{IJRRCopyright}{
    \lhead{\textsc{This paper is currently under review in The International Journal of Robotics Research} }
}

\def\volumeyear{2021}
\def\journalname{The International Journal of Robotics Research}
\hypersetup{pdftitle={Passive and Active Acoustic Sensing for Soft Pneumatic Actuators},
	pdfauthor={Vincent Wall and Gabriel Z{\"o}ller and Oliver Brock},
	pdfkeywords={Acoustic Sensing, Force and Tactile Sensing, Computational Sensor, Soft Sensors, Pneumatic Actuators, Soft Robotics
	pdfsubject=This paper is currently under review in The International Journal of Robotics Research}
}

\begin{document} 


\runninghead{Wall et al.}
\title{Passive and Active Acoustic Sensing \\ for Soft Pneumatic Actuators}
\author{Vincent Wall\affilnum{1,2}, Gabriel Z{\"o}ller\affilnum{1}  and Oliver Brock\affilnum{1,2}}

\affiliation{\affilnum{1}Robotics and Biology Laboratory, Technische Universit{\"a}t Berlin, Germany \affilnum{2}Science of Intelligence, Research Cluster of Excellence, Berlin, Germany}
\corrauth{Oliver Brock, Robotics and Biology Laboratory, Technische Universit{\"a}t Berlin, Marchstra{\ss}e 23, 10587 Berlin, Germany.}
\email{oliver.brock@tu-berlin.de}

\begin{abstract}
We propose a sensorization method for soft pneumatic actuators that uses an embedded microphone and speaker to measure different actuator properties. The physical state of the actuator determines the specific modulation of sound as it travels through the structure. Using simple machine learning, we create a computational sensor that infers the corresponding state from sound recordings. We demonstrate the acoustic sensor on a soft pneumatic continuum actuator and use it to measure contact locations, contact forces, object materials, actuator inflation, and actuator temperature. We show that the sensor is reliable (average classification rate for six contact locations of \SI{93}{\percent}), precise (mean spatial accuracy of \SI{3.7}{\milli\meter}), and robust against common disturbances like background noise. Finally, we compare different sounds and learning methods and achieve best results with \SI{20}{\milli\second} of white noise and a support vector classifier as the sensor model.
\end{abstract}

\keywords{Acoustic Sensing, Force and Tactile Sensing, Computational Sensor, Soft Sensors, Pneumatic Actuators, Soft Robotics}

\maketitle 
\thispagestyle{IJRRCopyright}

\section{Introduction}

We present sound-based sensing for soft robotic actuators. The underlying principle is simple: An object's physical state affects how sound is propagated through the object. The sound is modulated by the object's shape, its contacts with other objects, or forces exerted onto it. We show that a recording of this modulated sound permits the accurate reconstruction of the object's physical state. Acoustic sensing works with sound produced by interactions of the object with its environment (passive acoustic sensing) or by playing sounds from a small loudspeaker embedded into or attached to the object (active acoustic sensing). Using passive or active acoustic sensing, one might say that it is possible to "hear" the object's state.

Acoustic sensing is particularly well-suited for soft actuators and soft robots. Soft bodies change their state substantially as a result of actuation or compliant interactions with the environment. These changes have significant effects on the propagated sound, making the reconstruction of state from sound easier. Also, unlike most traditional sensing technologies, acoustic sensing does not constrain the actuator's morphology, thus permitting it to take full advantage of clever mechanical design and soft-material compliance. Furthermore, acoustic sensing eliminates the need to incorporate multiple special-purpose sensors (e.g. proprioception and contact sensors). We will show that acoustic sensing can emulate a variety of signal-specific sensors by recovering the different types of sensor information directly from sound. 

\begin{figure}
	\centering
	\includegraphics[width=\linewidth]{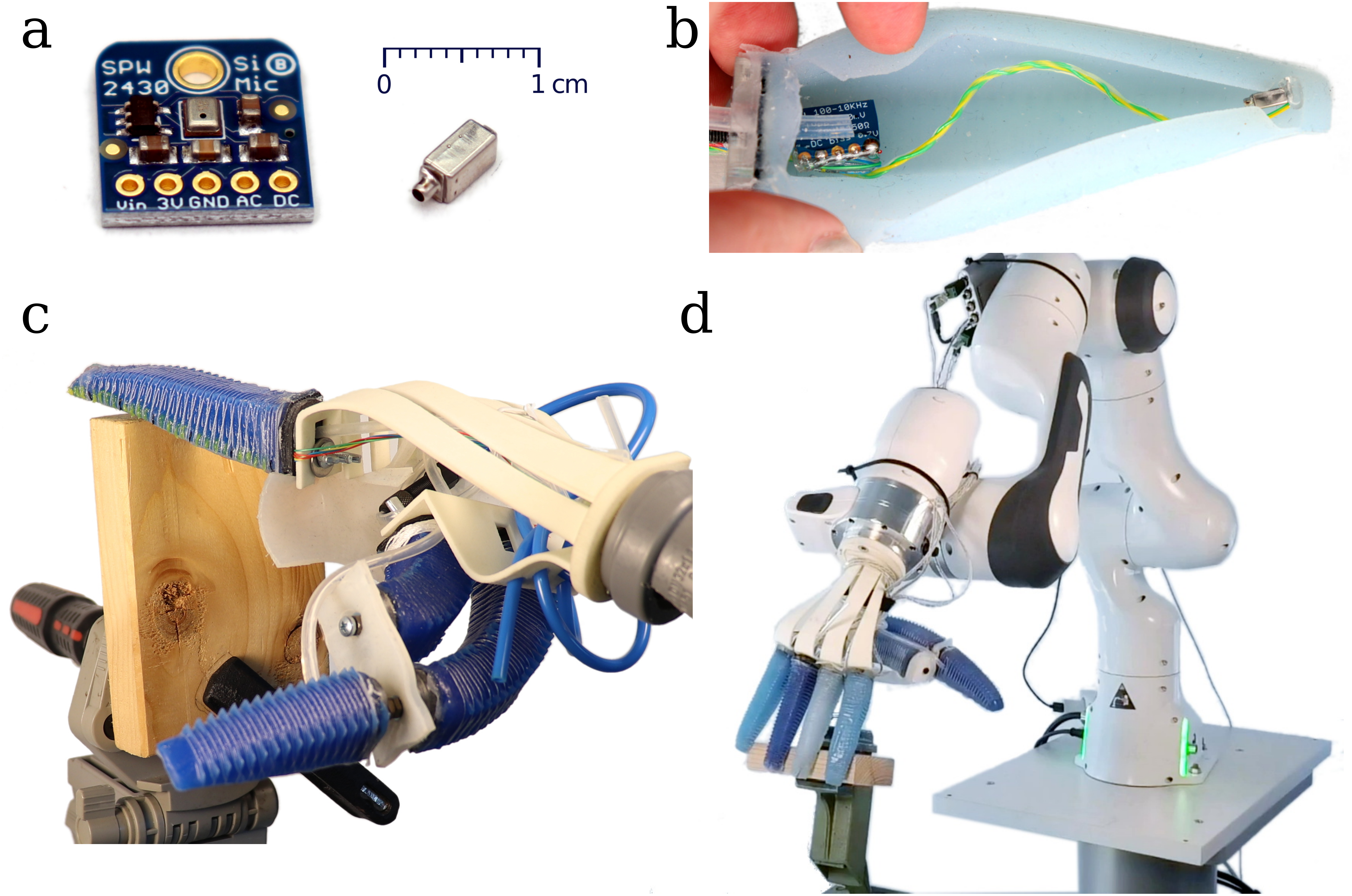}
	\caption{The acoustic sensor hardware inside a PneuFlex actuator: (a)~microphone~(left) and the speaker~(right), (b)~placement of the audio components inside the actuator, (c)~manual and (d)~automated data recording and test setup with the RBO~Hand~2}
	\label{fig:opener}
\end{figure}

In this paper, we provide a comprehensive description of how to deploy acoustic sensing in the context of soft robotics.  We also present an in-depth experimental evaluation of acoustic sensing with three types of experiments: First, we demonstrate the high accuracy and range of the sensor.  Then, we show the sensor's robustness against disturbances. Finally, we evaluate the effect of different sensor parameters.  The acoustic sensor achieves classification rates of up to \SI{100}{\percent} and a regression error as low as \SI{3.7}{\milli\meter}. The same sensor hardware measures different properties, like contact location and forces, as well as object materials and actuator inflation. At the same time, the acoustic sensor remains unaffected by background noises up to \SI{90}{\decibel}.

We believe that acoustic sensing is an extremely powerful, simple, and robust approach to sensing, particularly well-suited for soft robotics.  We hope that the comprehensive treatment of this sensing approach presented in this paper lays the foundation for further advancing acoustic sensing technology and for exploring novel applications.

The work in this article extends our previous publications on acoustic sensing~\citep{zoller_acoustic_2018, zoller_active_2020} with insights on the influence of several sensor parameters. We analyze different types of active sounds and the required sound volume. We compare different machine learning methods and evaluate the transferability of sensor models between different actuators. Furthermore, we add a novel demonstration of the sensor's capability to measure the actuator's temperature, as well as simultaneously measuring different actuator parameters from a single sound recording.

\section{Related Work} 
\label{sec:related_work}

Suitable sensorization for soft robotic actuators provides relevant measurements without negatively affecting compliance. We first review existing sensorization approaches for soft actuators and the degree to which they accomplish these goals, and then we survey prior approaches to acoustic sensing in robotics.

\subsection{Sensing for Soft Robotic Actuators}
\citet{yousef_tactile_2011}, \citet{kappassov_tactile_2015}, \citet{amjadi_stretchable_2016}, and \citet{wang_toward_2018} offer comprehensive reviews of sensing for soft actuators. Here, we summarize the most relevant sensorization approaches.

\emph{Strain sensors} measure the deformation of soft actuators through changes in sensor length~\citep{farrow_soft_2015, park_design_2012, vogt_design_2013, tapia_makesense_2020}.
Based on such measurements it is possible to infer information about the actuator's state including contact location and forces~\citep{wall_multi-task_2019}. Strain sensors are usually highly stretchable, reducing the negative effect on the overall compliance of the actuator. However, measurements contain information that is aggregated along the entire length of the sensor, which limits the spatial accuracy. Also, the most appropriate sensor placement is often task-specific but cannot easily be adapted.

\emph{Tactile sensors} for soft surfaces collect information about the intensity and location of contacts for a desired sensing area by measuring the deformation of the surface through various means~\citep{weigel_iskin:_2015, gerratt_elastomeric_2015, visentin_deformable_2016, buscher_flexible_2015, pannen_flexible_2022}. However, this usually requires the sensor to be placed directly at the point of measurement. So when sensorizing large surfaces, the additional material of the sensor hardware will negatively affect the compliance of the soft actuator.

\emph{Optical waveguides} measure the curvature of soft actuators by detecting changes in light intensity or frequency~\citep{to_soft_2018, zhao_optoelectronically_2016, galloway_fiber_2019}. When the waveguide is made from an elastic material, the influence on the actuator's compliance is minimal. But, like strain sensors, measurements aggregate over the whole length of the sensor, making it difficult to determine the exact origin of measurement. Alternatives, such as fiber bragg-grating~\citep{park_force_2007}, offer high precision, but they are much less compliant and require expensive read-out devices. 

\emph{Embedded cameras} use light to measure the deformation of soft actuators. By visually observing the backside of contact surfaces, for example measuring the light intensity or optical flow, highly detailed information about interactions can be obtained~\citep{nakao_finger_1990, ward-cherrier_tactip_2018, sferrazza_ground_2019}. And without the need for physical contact with the camera, the surface compliance remains unaffected. However, the sensor requires a line-of-sight to the point of measurement, which is problematic in structures with significant deformations, like continuum actuators which can have bend angles of over $180^\circ$.

In summary, current sensing approaches for soft actuators either provide limited detail due to the aggregation of measurements or significantly restrict the actuator's compliance. Furthermore, most sensors measure only a single actuator property. In contrast, our acoustic sensing approach has little effect on compliance and measures many different properties with high accuracy and at the same time.

\subsection{Acoustic Sensing in Robotics}

Sound is used to measure a wide range of different properties in both industry and research. For example, acoustic sensing has long been employed for fault detection in machines~\citep{takata_sound_1986}, railway infrastructure~\citep{lee_fault_2016}, and high-power insulators~\citep{park_acoustic_2017}. As another example, the term ``Distributed Acoustic Sensing'' describes the measurement of geomechanical strain in boreholes by observing minimal oscillations of fiber optic cables. Such ``acoustic antennas'' can measure rock displacements of less than \SI{1}{nm}~\citep{becker_distributed_2020}. In the medical field, sound has long been used for many different procedures, like ultrasound imaging~\citep{wells_ultrasound_2006}.
For this paper, however, we focus on previous approaches to acoustic sensing in robotic applications, the range of properties that can be measured, and its applicability to soft materials.

\subsubsection*{Sound Contains Diverse Information}
On the one hand, acoustic sensing can be used for \emph{exteroceptive} sensing, i.e.~measuring properties of other objects. Several approaches use sounds recorded while tapping, shaking, holding, etc., to recognize and classify objects~\citep{schenck_which_2014, sinapov_interactive_2011, kroemer_learning_2011, tomoaki_nakamura_multimodal_2007, richmond_active_2000, luo_knock-knock:_2017}. Recordings of sound have also been used to determine object materials~\citep{krotkov_robotic_1997}, surface properties~\citep{cuneyitoglu_ozkul_acoustic_2013}, and the distance from objects while grasping~\citep{jiang_seashell_2012}. Even features like the flow rate of granular material~\citep{clarke_learning_2018} or the ambient temperature~\citep{cai_active_2021} have been measured using sound.

On the other hand, acoustic sensing also provides \emph{proprioceptive} sensing capabilities, i.e.~measurements about the sensorized object itself. For example, novel touch interfaces were developed using sound/vibration recorded on rigid surfaces~\citep{collins_active_2009, harrison_tapsense_2011, liu_vibwrite:_2017, paradiso_passive_2002}. Furthermore, \cite{ono_touch_2013} recognized touch gestures on arbitrarily shaped objects using sound and could even measure the configuration of a structure built from Lego/Duplo blocks.

In this paper, we transfer these ideas from the domain of traditional ``hard'' robotics to the field of soft robots and show that the soft materials of pneumatic actuators are well suited for exteroceptive and proprioceptive measurements of object and actuator properties using acoustic sensing.

\subsubsection*{Sound Turns Whole Objects Into Sensors}

Because sound travels through structures, the recording location can be different from the origin of sound. This way, whole objects become sensors, as the structure transports information from anywhere on and in the object to the location of sound recording~\citep{ono_touch_2013, collins_active_2009, harrison_tapsense_2011, paradiso_passive_2002}. In contrast to previous approaches that used rigid structures, we apply this idea to a soft actuator. There it enables us to sense contact anywhere on the actuator while placing the microphone where it least influences the actuator's compliance.

\subsubsection*{Sound Is Suitable For Soft Structures}

Acoustic sensing has been shown to work well in ``soft'' structures. The small size of commercially available audio components makes it easy to embed acoustic sensing without affecting their compliance significantly. \citet{amento_sound_2002} used bone-conducted sound on human hands to recognize fingertip gestures. But without bones in pneumatic actuators, we cannot rely on them for sound modulation. In pneumatic actuators, \citet{takaki_acoustic_2019} used sound to measure the actuator's length by modeling the frequency response. However, for less symmetrical actuator shapes this is not feasible~\citep{rompf_entwicklung_2019}. \citet{mikogai_contact_2020} use sounds induced by a compressor to train a convolutional neural network that localizes contact on the supply tube of a pneumatic actuator. In our approach, we sensorize the actuator itself and use an embedded speaker directly inside its air chamber to actively choose which sound to play. Furthermore, in addition to contact locations, our sensor measures contact forces, object materials, inflation levels, and actuator temperature.

\section{Acoustic Sensorization}
\label{sec:method}

In this section, we first summarize the working principle of the acoustic sensor, before we present the implementation of the sensor's two core components: the physical sensor hardware and the computational component which extracts the desired measurements. Finally, we discuss two modes of sensing: passive and active.

\begin{figure}
	\centering
	{\flushleft \large a\\}	
	
	\includegraphics[width=\linewidth]{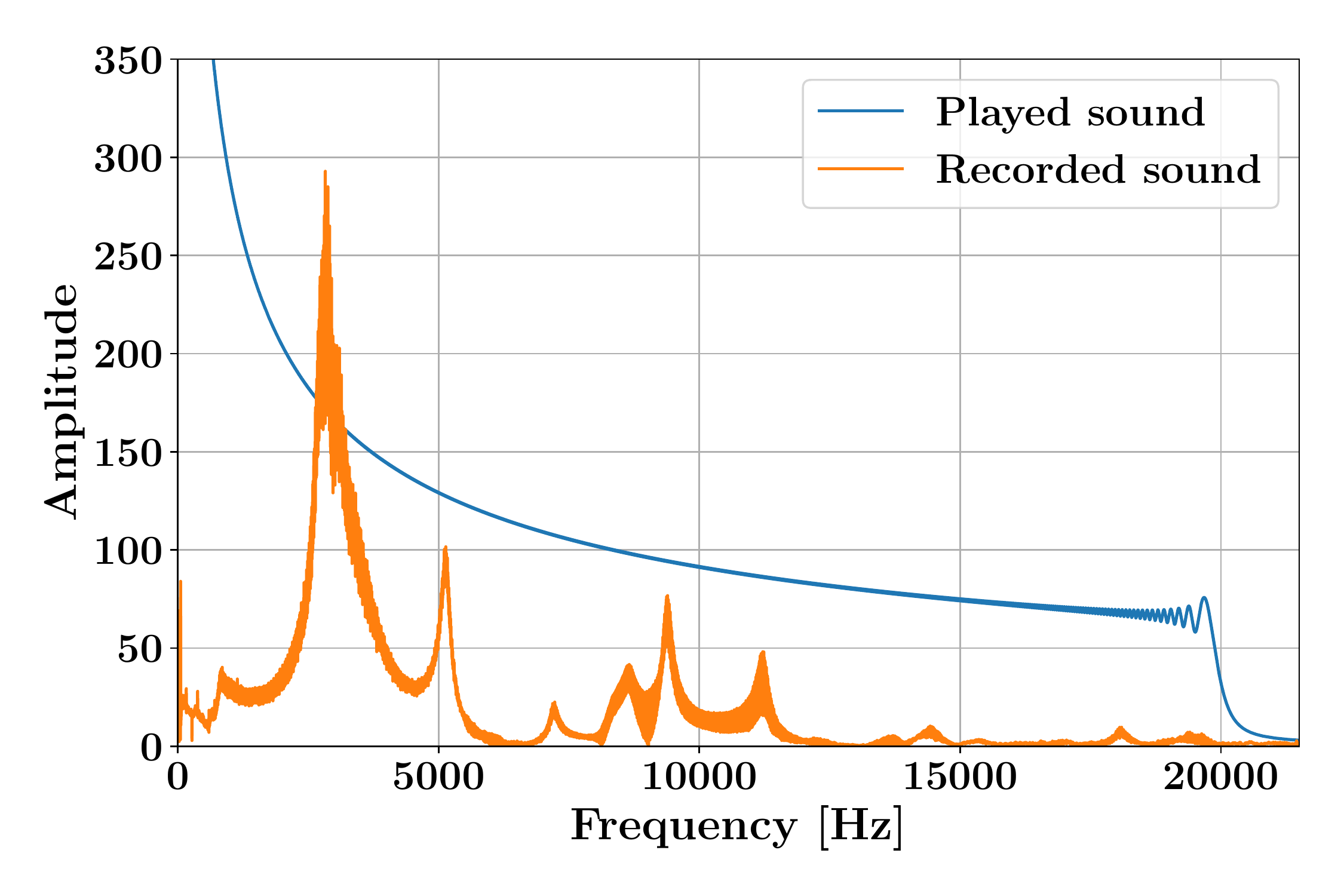}

	{\flushleft \large b\\}	
	\includegraphics[width=\linewidth]{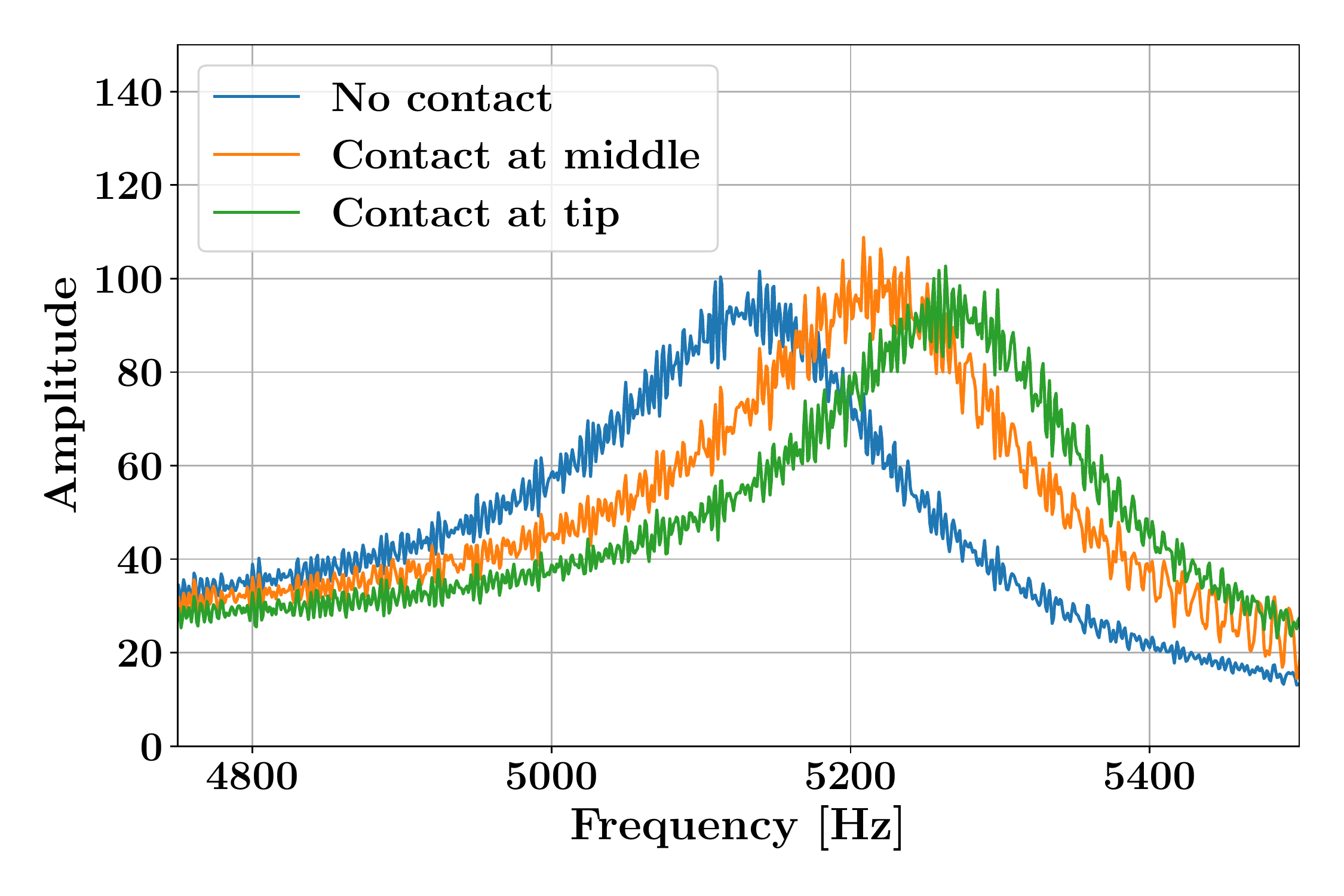}
	\caption{We record sound samples and convert them into frequency spectra. In the frequency domain, we observe a contact-dependent shift of the resonance peaks. (a) The frequency spectrum of a 'sweep' sound that serves as input to the speaker (blue) and the resulting spectrum of the sound as recorded by the microphone (orange) (b) The magnified peaks of spectra from different contact locations are noticeably shifted.}
	\label{fig:active_sounds}
\end{figure}

\subsection{Acoustic Sensing Principle} 
\label{sub:principle}

The key property of the acoustic sensing principle is the fact that sound that travels through an object gets modulated by it~\citep{cremer_structure-borne_2005}. And when certain properties of the object change, the modulation changes as well. Such properties include the object's shape and internal forces, but also interactions with the environment. Essentially, any change to the object which affects the physical transmission of sound waves. 
Additionally, the exact way that the sound is modulated is deterministic and often characteristic for one specific state of the object (where ``state'' means the space of all modulation-affecting object properties, like shape, interactions, etc.) Consequently, it is possible to \emph{infer} the object state from a given sound modulation. So by recording sound within the object, we can observe a given modulation and use that to determine the current state of the object.

But this mapping between object state and sound modulation is complex. Many object properties will affect it. The creation of an \emph{analytic} model will only work for very simple relations, such as the change of the resonance frequency when the cavity size is changed~\citep{takaki_acoustic_2019}. But other effects, like changes due to deformation or contact, are too complicated to model accurately, especially for soft material objects with asymmetric shapes~\citep{rompf_entwicklung_2019}. 
Instead, we propose to use machine learning techniques to solve the inverse problem. Using data-driven supervised learning methods, we create \emph{empirical} models to map from recorded sounds to measured object properties.

Such a combination of a \emph{physical} sensor signal and the subsequent extraction of the relevant measurement through \emph{calculation} is called a ``computational sensor''~\citep{van_der_spiegel_computational_1996}. With it, the measured property no longer depends on the specific sensor hardware. Instead, a more generic sensor signal, e.g.~a microphone recording, is converted into the desired sensor measurement via a computational interpretation of the data. Such a computational sensor can \emph{emulate} a wide range of different sensor types, like a contact sensor, force sensor, inflation, and even temperature, all while using the same, simple sensor hardware, which in our case is an embedded microphone.

The whole acoustic sensing principle is based on the sound modulation in objects. As a result, it can be applied to any sound-conducting structure. As such, it has been used to sensorize objects as diverse as Lego blocks~\citep{ono_touch_2013}, the human hand~\citep{amento_sound_2002}, or storefront windows~\citep{paradiso_passive_2002}. Many other applications are possible.
In this paper, we demonstrate the application of the acoustic sensing principle to the domain of soft robotic actuators. The unintrusive design, as well as the large range of measurable properties, make it a highly versatile sensorization approach.

\subsection{Design and Fabrication of the Sensor Hardware} 
\label{sub:sensorized_actuator}

The physical component of the acoustic sensor consists of a microphone and speaker that must be placed in or on the actuator. We pursue two main goals: First, we want to minimize the detrimental effect on compliance, and second, we want to allow for substantial sound modulation when sound propagates inside the actuator.

We sensorize a PneuFlex actuator~\citep{deimel_compliant_2013}. This pneumatic continuum actuator is made of highly-flexible silicone rubber with an air chamber that spans the full length of the finger. The audio components are embedded into the actuator during fabrication before the air chamber is sealed. To maximize the travel distance of the sound, we place the speaker and the microphone at opposite ends of the actuator's air chamber. The microphone is placed at the base of the actuator because there is more space and compliance is less affected. Both components are attached to the actuator using silicone adhesive (Sil-Poxy). The speaker cables are guided through the air chamber with sufficient slack to allow for actuator deformations (Fig.~\ref{fig:opener}b). Cables exit the air chamber at the base of the actuator, where a thin coating of silicone on the cables ensures air-tightness.

We embed a MEMS (micro-electro-mechanical system) condenser microphone (Adafruit SPW2430) into the actuator. It has a wide frequency range (linear response between \SI{100}{\hertz} and \SI{10}{\kilo\hertz}) and a small form factor (see Fig.~\ref{fig:opener}a). The breakout board also includes a convenient on-board \SI{3}{V} power regulator. To further reduce its size, we file off any unnecessary parts of the board (Fig~\ref{fig:opener}b). 

We also embed a balanced armature speaker (Knowles RAB-32063-000) into the actuator. It has a comparable frequency range (\SI{80}{\hertz} to \SI{10.8}{\kilo\hertz}) and is small in size.
A USB audio interface (MAYA44 USB+) drives both audio components at a sample rate of \SI{48}{\kilo\hertz} with \SI{32}{bit} precision.

\subsection{Creating the Sensor Model} 
\label{sub:training_sensor_model}

The computational component of the sensor transforms the recorded sound signal into the sensor measurement of the computational acoustic sensor. This process consists of three parts: data pre-processing, training of the sensor model, and evaluation of the model.

First, we pre-process the recorded sound samples by trimming them to the same length and converting them into the frequency domain using a Discrete Fourier Transform. The frequency representation exposes the resonance effects of the air chamber. We use only the real-valued amplitude spectrum and disregard the complex-valued phase spectrum. This simplifies the signal as temporal patterns are ignored. In the future, we will use the phase spectrum to extract additional information. The resulting feature vector contains the amplitude of each frequency in the signal from \SI{1}{\hertz} to \SI{24}{\kilo\hertz}. Even though this vector can be quite large (24k values/sec), we observed no need for down-sampling in our experiments. Figure~\ref{fig:active_sounds}a shows an example sound recording converted into the frequency spectrum. 

If we had an analytical model of the actuator's acoustic behavior, we could now directly calculate the origin of the observed modulation. However, as~\cite{rompf_entwicklung_2019} has found, the complex shape and material of soft actuators make it infeasible to create reliable analytical models. Instead, we use supervised learning to train an empirical sensor model. Figure~\ref{fig:active_sounds}b) shows an example of the frequency shift that appears in sound signals from different contact states. This is the result of the state-dependent sound modulation of the actuator and is highly repeatable. Hence, we can train the sensor model to recognize these state-dependent patterns. We use a simple k-nearest neighbor (KNN) classifier. It is extremely fast to ``train,'' because it simply remembers the training samples, and requires only a small number of samples per class. In our experiments, we use between~5 and~25 samples per class.

In Section~\ref{sub:ml_methods} we analyze if more complex learning methods lead to increased sensor accuracy. However, for the remainder of the paper, we will use the KNN predictor as a lower bound. We use the default implementation of the KNN classifier from the scikit-learn library~\citep{pedregosa_scikit-learn:_2011}: $n=5$ neighbors and a Euclidean distance metric.

Finally, we evaluate the sensor model on a separate test set. We split each data set 3:2 into training and test data while maintaining equal class distributions. The results are given as the average classification rate (ACR) across all classes, evaluated on the previously unseen test data.\footnote{All code and data sets are available at this link: \url{http://dx.doi.org/10.14279/depositonce-11059.2} \\
Additionally, we provide an ``Acoustic Sensing Starter Kit'' to easily create your own acoustic sensor: \url{http://bit.ly/AcousticSensing}
}.

The trained sensor model together with the physical hardware described in the previous section comprises the proposed computational acoustic sensor.

\subsection{Two Sensing Modes: Passive and Active} 
\label{sub:passive_active}

The proposed computational acoustic sensor requires sound to work. We consider two different sound sources: In \emph{passive} sensing, the sound comes from the environment, either from external noises or from an impact of the actuator with its environment. As we will show, even light contacts create distinct sounds inside the air chamber. In \emph{active} sensing, the sound is created by the embedded speaker. This way, we have full control over the sound that serves as input to our sensor. We can generate it at any time to probe the system state, and we can select it based on the sensing task.

\begin{figure*}
	\centering
	\begin{minipage}[t]{0.2\textwidth}
		\large{a} \hfilll \small{Contact Locations} \hfilll \\
		\includegraphics[width=\textwidth]{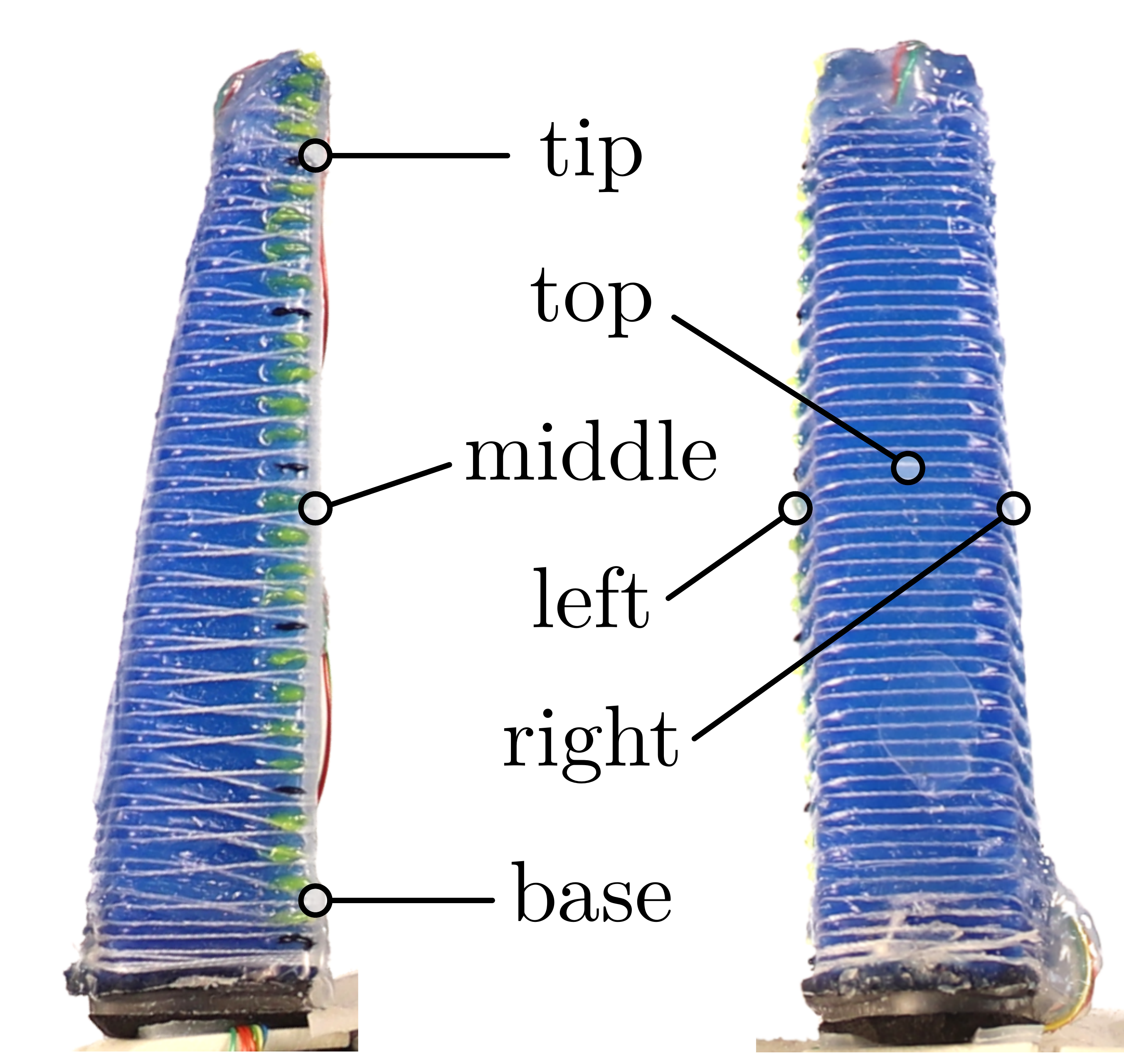}
	\end{minipage}
	\begin{minipage}[t]{0.26\textwidth}
		\large{b} \hfilll \small{Passive} \hfilll \\
		\includegraphics[width=\textwidth,trim={0.5cm, 1.2cm, 0.2cm, 2.6cm},clip]{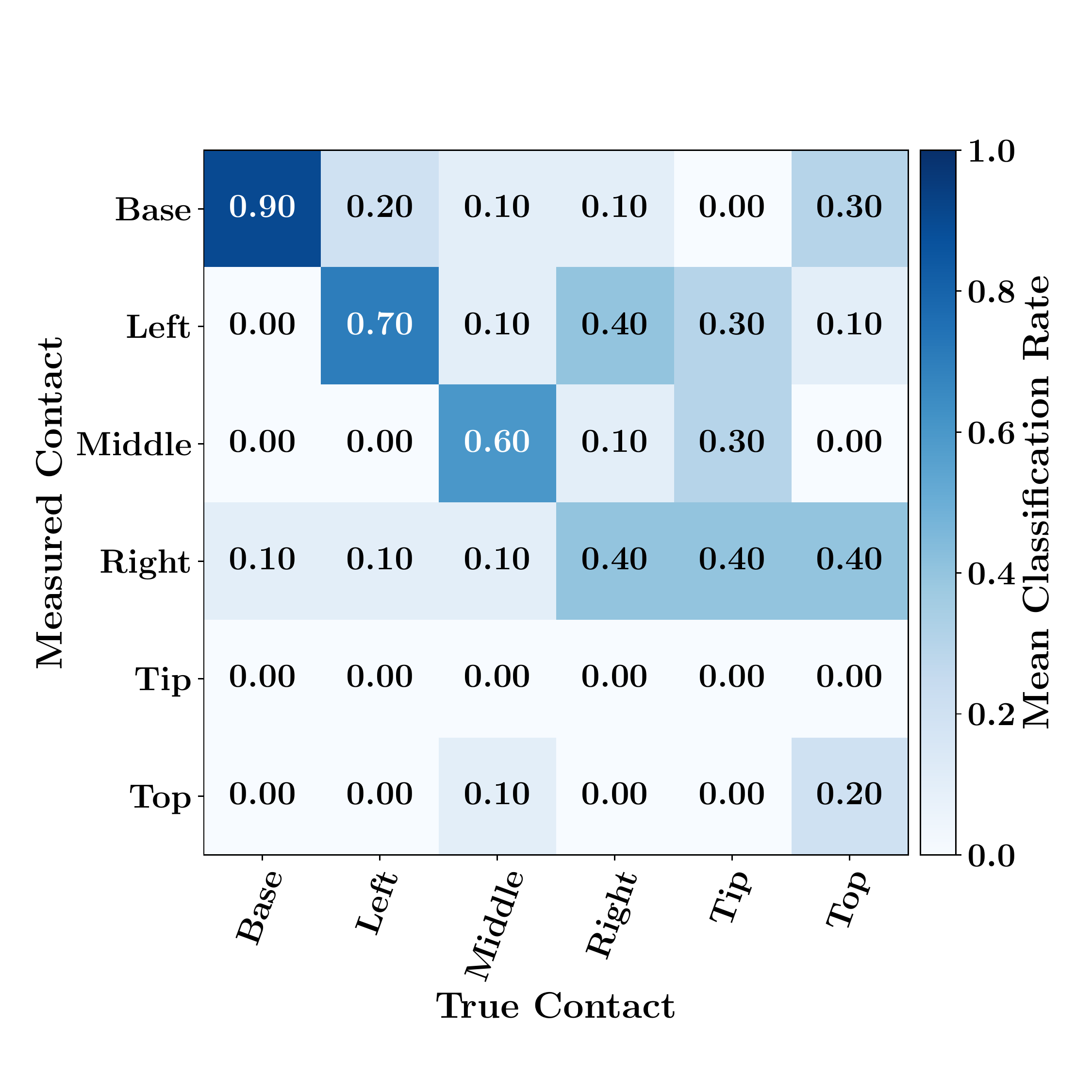}
	\end{minipage}
	\begin{minipage}[t]{0.26\textwidth}
		\large{c} \hfilll \small{Dynamic} \hfilll \\
		\includegraphics[width=\textwidth,trim={0.5cm, 1.2cm, 0.2cm, 2.6cm},clip]{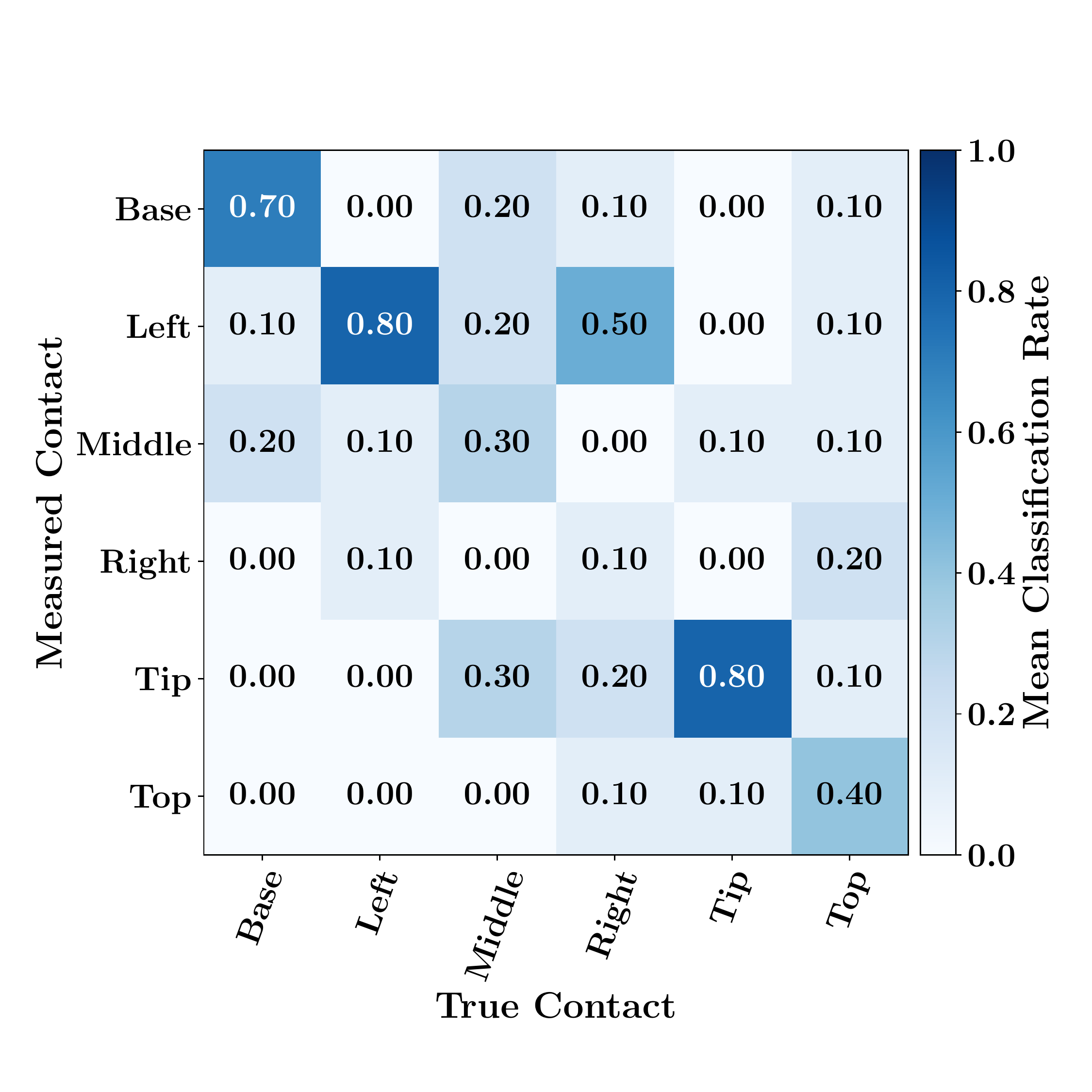}
	\end{minipage}
	\begin{minipage}[t]{0.26\textwidth}
		\large{d} \hfilll \small{Active} \hfilll \\
		\includegraphics[width=\textwidth,trim={0.5cm, 1.2cm, 0.2cm, 2.6cm},clip]{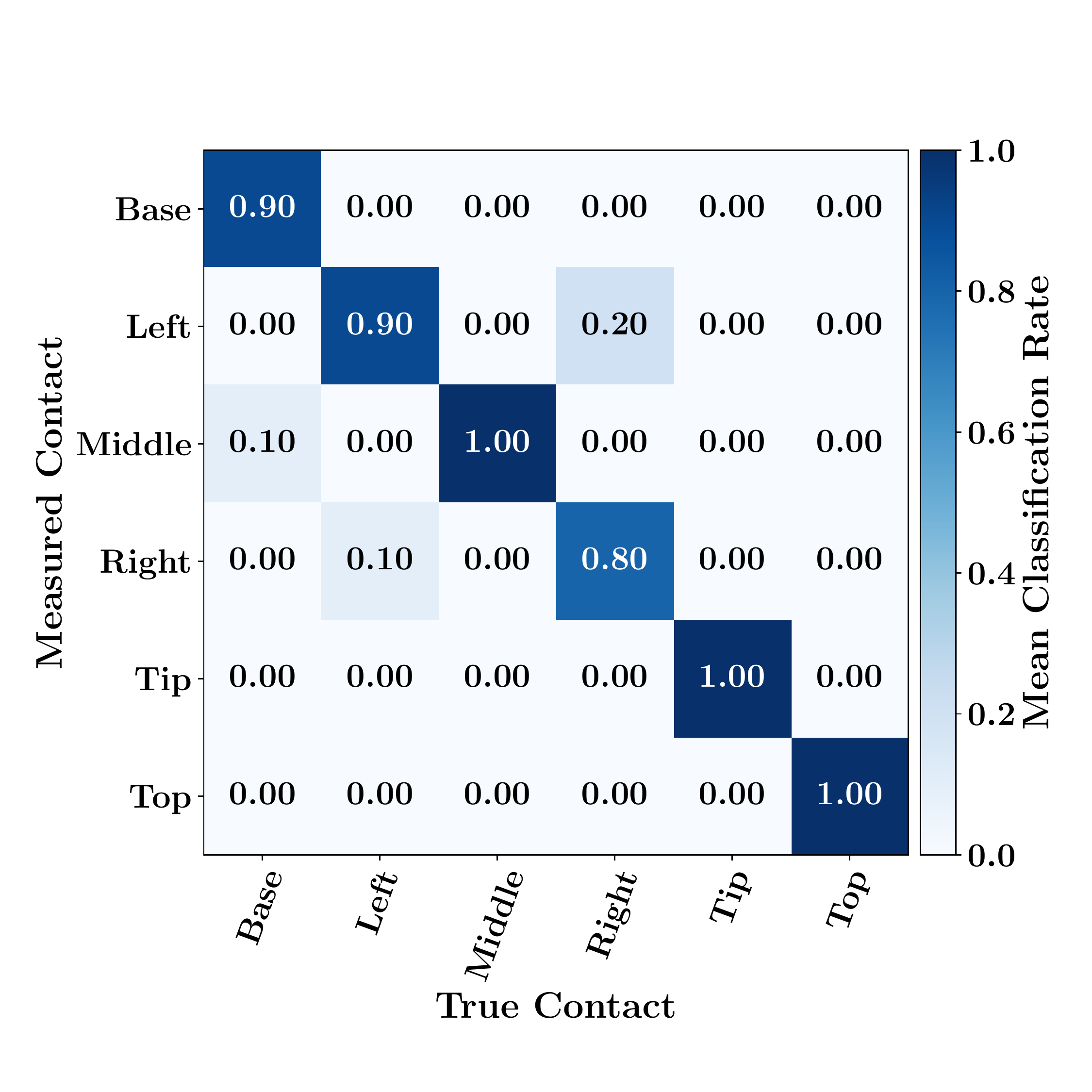}
	\end{minipage}

	\caption{Contact location sensing: \textbf{(a)}~We define six contact locations distributed across the PneuFlex actuator. \textbf{(b)}~Confusion matrix for passive sensing with environmental noise, \textbf{(c)}~dynamic sensing with contact sounds, \textbf{(d)}~and active sound played by the embedded speaker. The active sensor achieves the best results with an average classification rate of \SI{93}{\percent}.
	}
	\label{fig:exp_contact_loc}
\end{figure*}

\section{Experimental Validation of the Emulated \\Sensor Types}
\label{sec:experiment}

In this section, we demonstrate the effectiveness of the acoustic sensor by presenting several proprioceptive and exteroceptive sensing examples. These experiments highlight the diversity and accuracy achievable with acoustic sensing.
In the sections afterward, we then analyze the sensor's robustness against common disturbances (Section~\ref{sec:disturbances}) and the influence of different sensor parameters (Section~\ref{sec:method_parameters}), before discussing the limitations of the approach (Section~\ref{sec:limitations}).

\subsection{Acquisition of Training and Test Data} 
\label{sub:recording_setups}

Before we can create sensor models, we must obtain labeled sound samples from the different classes of actuator states.  We attach the sensorized actuator as the index finger to an RBO~Hand~2 and record sounds using the embedded microphone. For each data point, the actuator is brought into the corresponding contact state and the active and/or passive sound is recorded. The recording order of the different classes is randomized to eliminate any temporal effects. 

Figures~\ref{fig:opener}c-d show the two recording setups we use: In the ``manual'' setup, we attach a handle to the hand for a human operator, who can place the actuator into any desired configuration, without being restricted by a limiting robot workspace.  In the ``automated'' setup, we mount the hand on a 7-DoF Panda robot arm and use pre-recorded robot poses to bring the actuator into contact with the object. While this setup takes more effort to prepare, it is more accurate and simplifies the recording of large data sets.

To ensure that the speaker sound for the active sensor contains all relevant frequencies, even though we have no reliable acoustic model for the actuator~\citep{rompf_entwicklung_2019}, we chose to generate sounds that span the complete range of frequencies the microphone can record. We use two different sounds: a  logarithmic frequency sweep from \SI{20}{\hertz} to \SI{20}{\kilo\hertz}, and random white noise. (Additional sounds are evaluated in section~\ref{sub:other_sounds}).  In the sweep, each frequency appears sequentially, which may lead to more distinct resonance effects. The white noise contains all frequencies simultaneously, which simplifies sample alignment and reduction of sample length.  We generate the sounds using the LibROSA Python library~\citep{mcfee_librosa_2015}. For synchronized playback and recording of sounds, we use the open-source software QjackCtl\footnote{\url{https://qjackctl.sourceforge.io/}} and Zita-Jacktools\footnote{\url{http://kokkinizita.linuxaudio.org/}}.

\subsection{Sensing Contact Locations With \SI{93}{\percent} Classification Rate}
\label{sub:contact_location}

We start with an experiment that demonstrates the acoustic sensor's core functionality: reliably measuring relevant actuator states from sound recordings. For this, we train the sensor to differentiate between six contact locations distributed across the hull of the actuator. Additionally, this experiment proves that a single microphone is sufficient to sense contact anywhere on the actuator. 
Such contact measurements provide valuable feedback for any robotic application that uses tactile feedback, e.g.~to reduce the uncertainty during motion planning~\citep{pall_contingent_2018} or to further improve robustness during in-hand manipulation~\citep{bhatt_surprisingly_2021}.

We use the manual recording setup for this experiment. We define six contact location categories on the actuator: base, middle, tip, left, right, and top (see Fig.~\ref{fig:exp_contact_loc}a). We record separate data sets for active and passive acoustic sensing. We further distinguish between completely passive sounds (only from the environment) and dynamic sounds (no active sound, but the actuator is tapped against the object once, which creates a contact sound). The active sound is a \SI{1}{\second} sweep. Each data set consists of 150~samples (six contact locations $\times$ 25 repeats). The KNN-classifier is created from the converted training data set (see Section~\ref{sub:training_sensor_model}). 

Using the trained KNN-classifier on the test set provides prediction results summarized in the confusion matrices in Figures~\ref{fig:exp_contact_loc}b-d for passive, dynamic, and active acoustic sensing, respectively. The results are normalized, showing the ratio of predictions per class. High values on the diagonal represent a good classification. That is the case for the \emph{active} sensor, with an average classification rate (ACR) of \SI{93}{\percent}. Only four of the 60 test samples were misclassified.

The passive and dynamic sensors achieve slightly lower average classification rates, \SI{47}{\percent} and \SI{52}{\percent}, respectively\footnote{By using a support vector classification instead of the KNN, the dynamic sensing classification rate is improved to \SI{74.7}{\percent} (see~\citet{zoller_acoustic_2018}). We consider the simpler KNN classifier as a lower bound. See Sec.~\ref{sub:ml_methods} for a comparison of different learning methods.}. Interestingly, however, even the passive sensor using only environmental noises performs significantly above the baseline of random chance (\SI{17}{\percent}).

This shows that in all three sensing modes---passive, dynamic, and active---the acoustic sensor can pick up on small regularities in the recorded sounds and use these to extract the relevant information about the location of contact.

\subsection{Sensing With a Spatial Accuracy of 3.7\,mm}
\label{sub:accuracy}

We evaluate the achievable spatial accuracy of our acoustic sensing setup by performing a contact location regression along the actuator. A low average error will indicate a high spatial accuracy of the sensor. 

\begin{figure}  
	\centering
	{\flushleft \large a\\}%
	\includegraphics[width=0.9\linewidth]{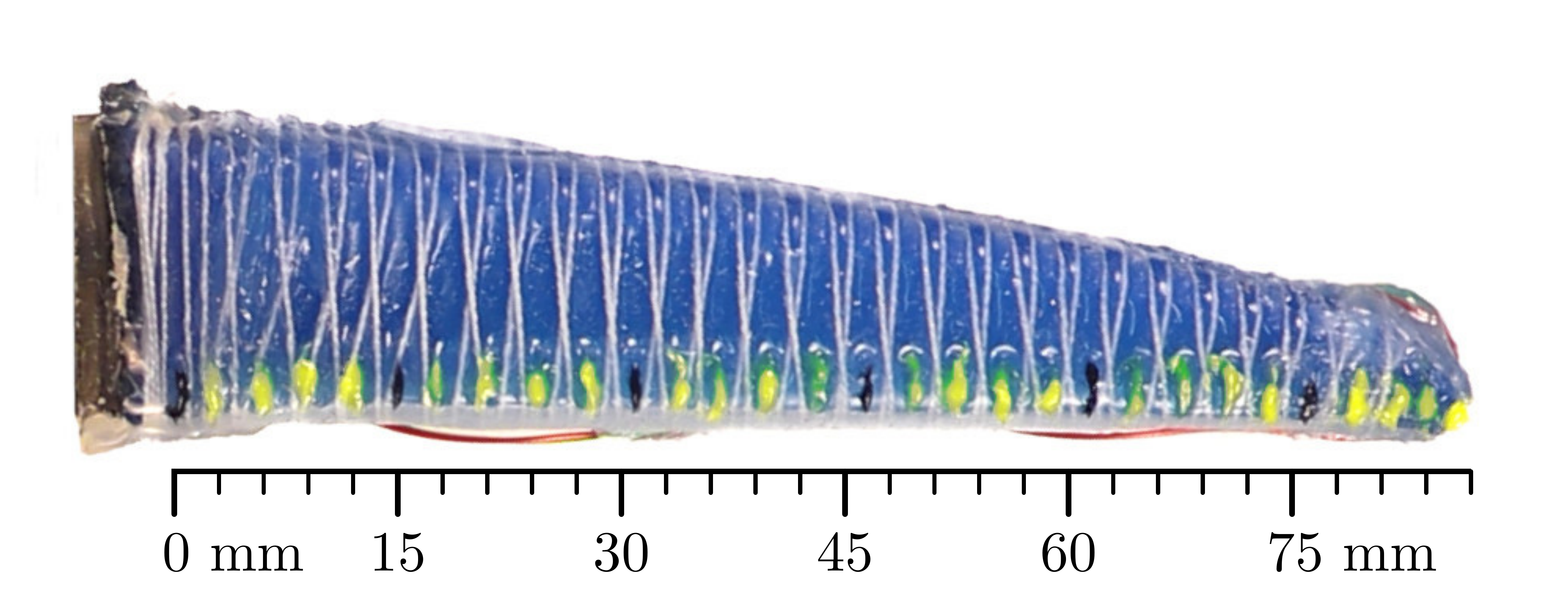}%
	{\flushleft \large b\\}%
	\includegraphics[width=1\linewidth]{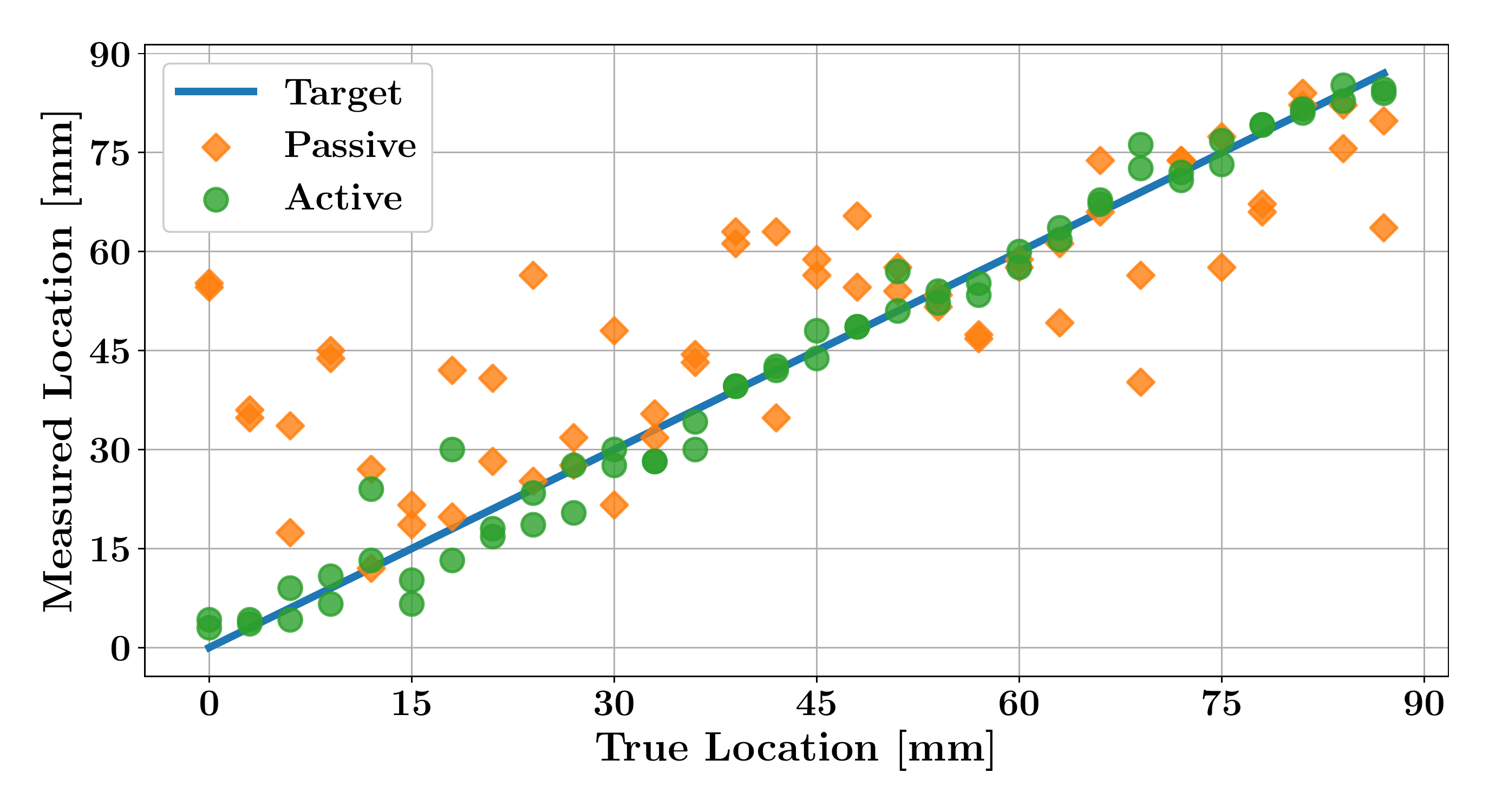}%

	\caption{Contact location accuracy evaluation: \textbf{(a)}~We record samples for 30 contact points along the actuator. \textbf{(b)}~The passive sensing measurements (orange squares) are not very accurate (RMSE: \SI{18.0}{\milli\meter}). The measurements with the active sensor (green circles) show only small deviations from the target (blue line) which demonstrates high accuracy (RMSE: \SI{3.7}{\milli\meter}).
	}
	\label{fig:exp_contact_accuracy}
\end{figure}

We mark 30~locations along the palmar side of the actuator, \SI{3}{\milli\meter} apart (see Fig.~\ref{fig:exp_contact_accuracy}a). Using the manual recording setup, we record each location five times with separate  data sets for passive and active sensing. The active sound is a \SI{1}{\second} sweep. For the model, we use a KNN-regressor ($n=5$ neighbors, uniform weights, Euclidean distance metric). This allows the prediction of the contact location on a continuous scale. 

In Figure~\ref{fig:exp_contact_accuracy}b, the true and predicted contact locations of the 60 test samples are compared. The passive predictions deviate noticeably from the target line with a root-mean-square error (RMSE) of \SI{18.0}{\milli\meter}. The passive sounds from vibrations and noises appear too limited in their expressiveness for an accurate prediction of the contact location. In the case of active sensing, however, the predictions are very close to the target with an RMSE of only \SI{3.7}{\milli\meter}. This demonstrates the high spatial accuracy that is achievable with only a single microphone and speaker embedded into the actuator. 

\subsection{Sensing Contact Forces With \SI{98}{\percent} Classification Rate}
\label{sub:forces}

We now demonstrate the potential of acoustic sensing to emulate different types of sensors. Using the same sensor hardware, we change the sensor model to measure a different actuator property: the contact force. 
Proprioceptive sensing of contact forces is a useful tool for soft robotics because the complex deformation during interactions makes it difficult to use other force sensors. And it is especially useful in applications that require a soft touch, e.g.~for human-robot interaction~\citep{knoop_handshakiness:_2017} or pick-and-place of delicate fruits and vegetables~\citep{mnyusiwalla_bin-picking_2020}.

We use the active sensor in the automated recording setup. The active sound is \SI{20}{\milli\second} of white noise. Additionally, we mount a force/torque sensor in between hand and wrist. We collect data for three forces: \SI{0.5}{N} (light contact), \SI{1.5}{N} (medium contact), and \SI{3}{N} (strong contact). A single contact location ('middle') is used. The actuator makes contact with the wooden object from three sides. We record 225 samples in total and create a KNN classifier from the training data.

\begin{figure*}  
	\centering
	\begin{minipage}[t]{0.28\textwidth}
		\large{a} \hfilll \small{Contact Force} \hfilll \\
		\includegraphics[width=\textwidth,trim={0.3cm, 0cm, 0.3cm, 0cm},clip]{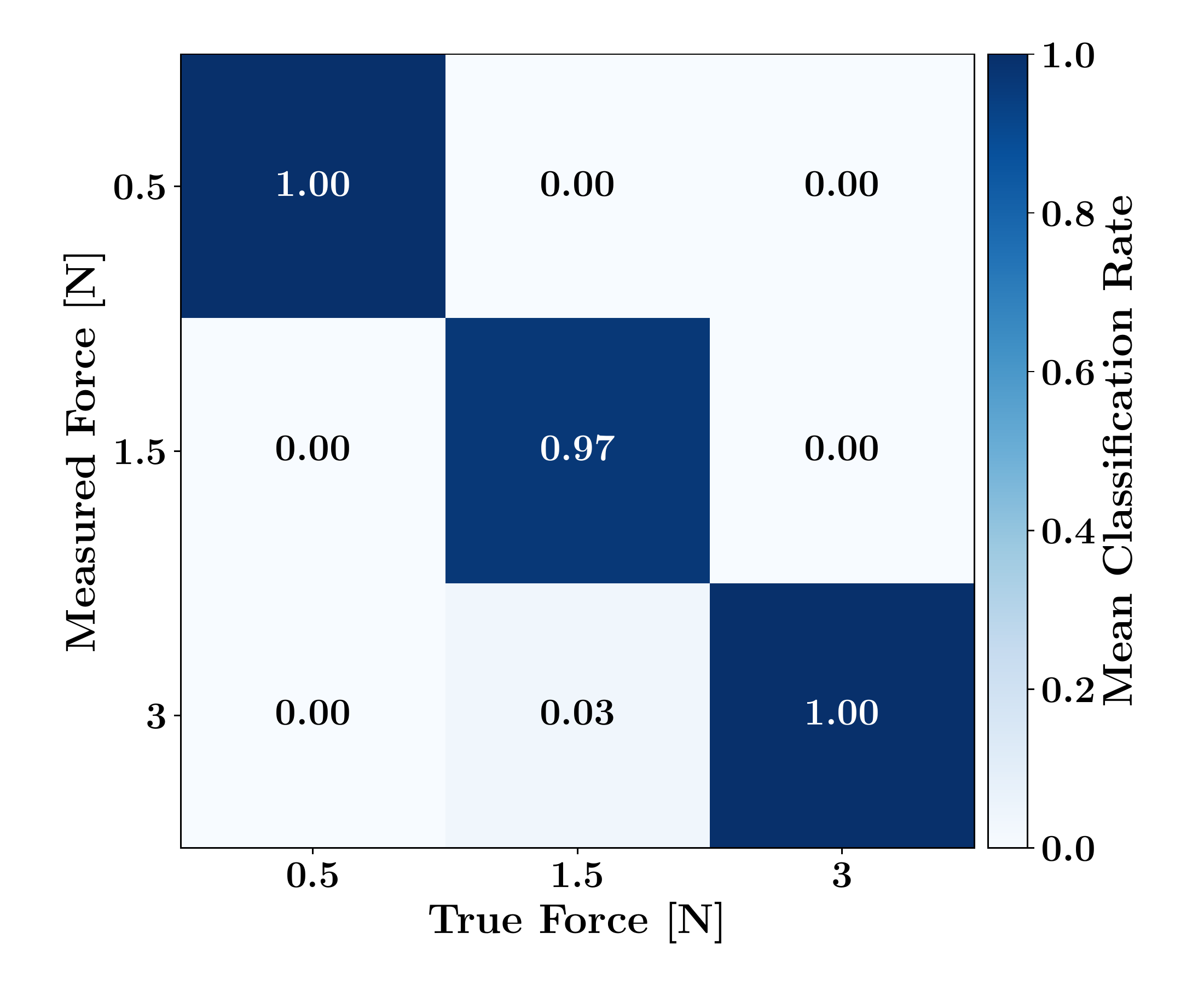}
	\end{minipage}
	\begin{minipage}[t]{0.28\textwidth}
		\large{b} \hfilll \small{Object Material} \hfilll \\
		\hfilll
		\includegraphics[width=0.9\textwidth, trim=1.5cm 0.6cm 2cm 0.25cm, clip]{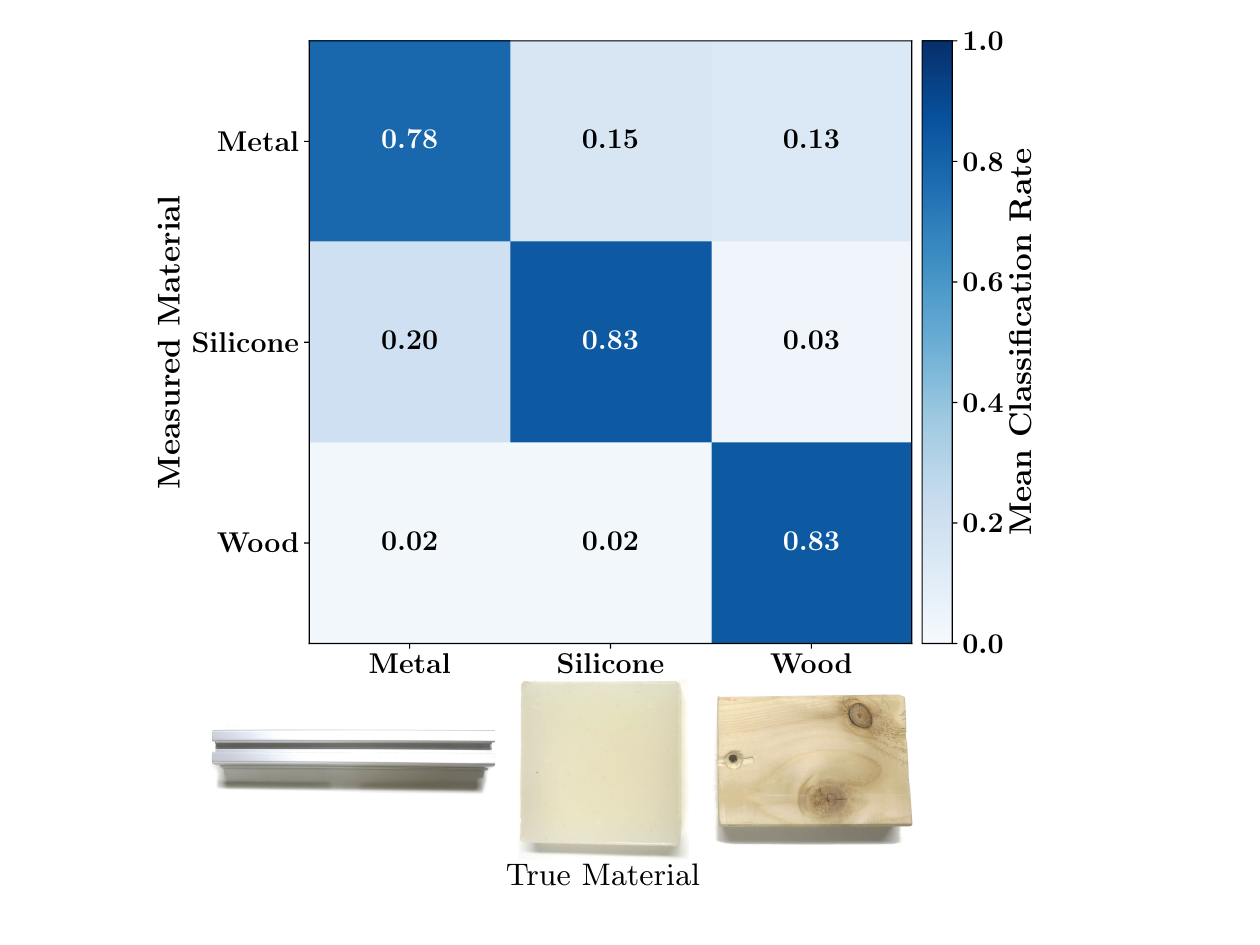}
		\hfilll
		\vspace{-0.5cm}
	\end{minipage}
	\begin{minipage}[t]{0.4\textwidth}
		\large{c} \hfilll \small{Actuator Temperature} \hfilll \vspace{0.2cm} \\
		\includegraphics[width=\textwidth,trim={0.7cm, 0.5cm, 0.7cm, 0.5cm},clip]{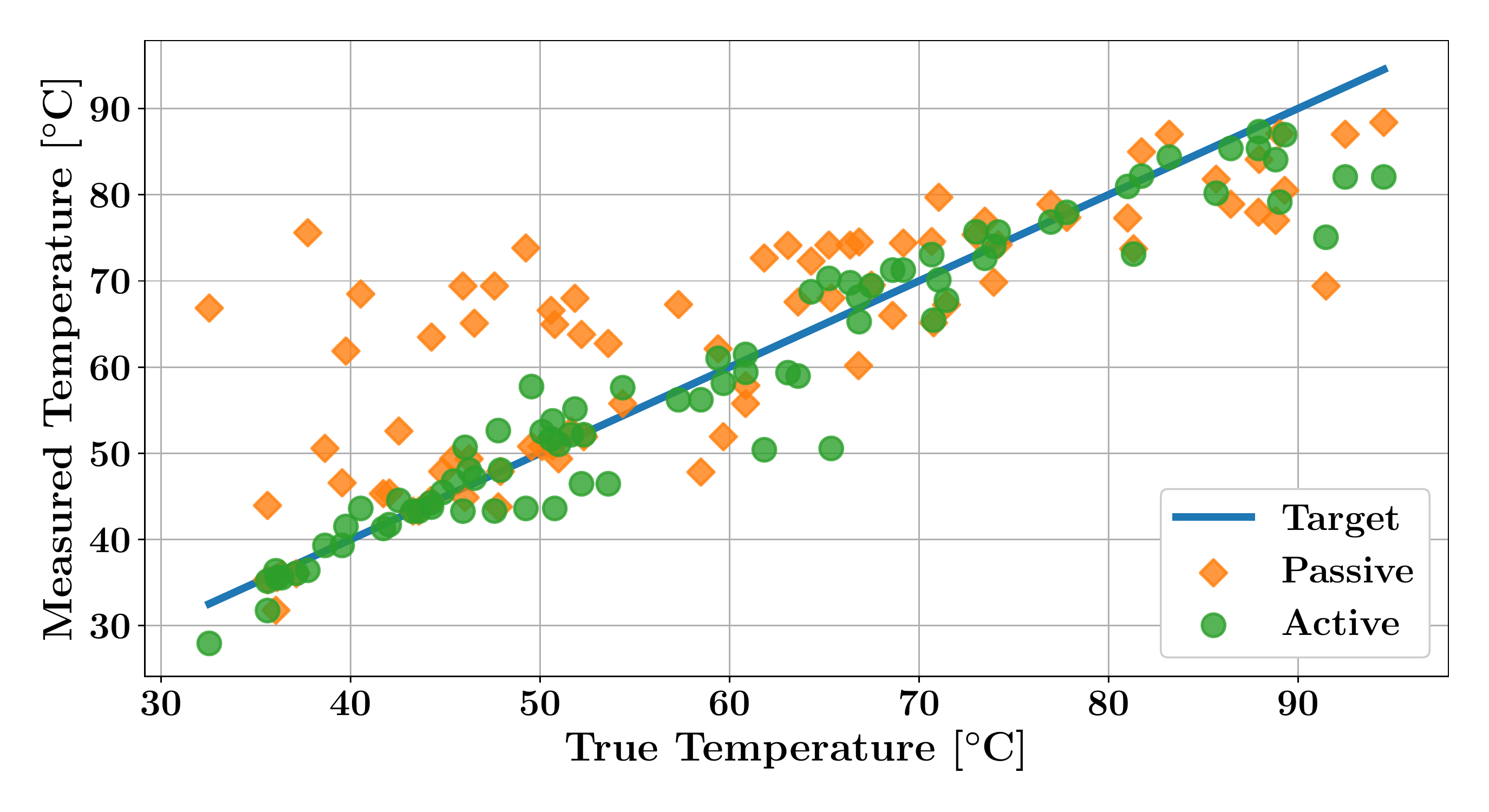}
	\end{minipage}

	\caption{Acoustic Sensing of different properties \textbf{(a)}~Contact force sensing: Three contact force classes are measured reliably (ACR \SI{98}{\percent}). \textbf{(b)}~Object material sensing: Three object materials are distinguished well (ACR \SI{82}{\percent}). \textbf{(c)}~Temperature sensing: The passive sensor (orange squares) measures the rough region of the actuator temperature (RMSE \SI{10.9}{\celsius}). The active acoustic sensor measurements (green circles) show only small deviations from the target (blue line), demonstrating high temperature accuracy (RMSE \SI{4.5}{\celsius}). These results illustrate the high versatility of the acoustic sensing approach, using only audio components and computation to measure a wide range of properties.
	}
	\label{fig:exps_force_material_temperature}
\end{figure*}

Figure~\ref{fig:exps_force_material_temperature}a shows the confusion matrix for active sensing of the contact force. The high values on the diagonal show that for almost every test sample the contact force was measured correctly with a classification rate of \SI{98}{\percent}. This demonstrates that besides the location of a contact, also the force of contact results in distinctive modulation of sound by the actuator. With our data-driven training approach, we can create sensor models that recognize the force-specific patterns in the sound's frequency spectrum.

\subsection{Sensing Object Material With \SI{82}{\percent} Classification Rate}
\label{sub:material}

Next, we evaluate the sensor's ability to measure the material of the touched object. Even though this property is not directly related to the actuator itself, it nonetheless becomes observable through contact. 
During contact, the acoustic properties of the touched object also affect the modulation of sound within the actuator. This is similar to the sound difference that is observable when objects of different material are tapped or struck~\citep{krotkov_robotic_1997}. We use this to determine the object material using the acoustic sensor.
In an application, such measurements could help to characterize grasped objects during exploration tasks~\citep{kroemer_learning_2011, sinapov_interactive_2011}.

We record data for contact with three different objects: a block of wood, a block of silicone, and an aluminum rod. All other properties (contact locations, contact force, etc.) are kept identical. 
We record six static contact locations for each object, using the manual recording setup and a \SI{1}{\second} sweep as active sound. Each contact location is recorded five times in random order, totaling 450 recordings. From the training data, we create a support vector machine (SVM)-classifier with the following parameters: linear kernel, $C=100$, $\gamma=$'scale'. For a comparison of different machine learning algorithms see Section~\ref{sub:ml_methods}.

Figure~\ref{fig:exps_force_material_temperature}b shows the confusion matrix of the material sensing. The three materials are recognized with high reliability with a mean classification rate of \SI{81.6}{\percent}. While this result is slightly lower than the contact location and contact force classifications, it nonetheless shows the impressive ability of the acoustic sensor to perform exteroceptive sensing of objects in contact with the actuator, by using sound recorded within the actuator's air chamber. 

\subsection{Sensing Temperature With a Mean Accuracy of \SI{4.5}{\celsius}}
\label{sub:temperature_sensing}

Temperature is another measurable property, whose connection to the acoustic modulation within the actuator might not be immediately obvious. But both microphone and speaker, as well as the air chamber and the surrounding silicone material, appear to be affected by changes in temperature. In this experiment, we evaluate the accuracy with which the acoustic sensor can measure temperature.

We place the actuator in an electric oven and gradually adjust the temperature between \SI{20}{\celsius} and \SI{95}{\celsius}. Every \SI{15}{\second} we record samples for both passive and active sensing. In the passive case, the only sound comes from the oven fan. For the active case, we use the \SI{1}{\second} white noise signal. We use an infrared thermometer to record the ground truth temperature information. For each case, we record 250 samples, which are shuffled and randomly split into two-thirds training and one-third test data. For the sensor model, we use a simple KNN-regressor ($n=5$ neighbors, uniform weights, Euclidean distance metric).

Figure~\ref{fig:exps_force_material_temperature}c shows the true and predicted temperature measurements. The \emph{passive} acoustic sensor achieves a root-mean-square error (RMSE) of \SI{10.9}{\celsius}, while the \emph{active} acoustic sensor achieves an RMSE of \SI{4.5}{\celsius}. This demonstrates that some component in the complex structure of the actuator is affected by the temperature change in a way, that the recorded sound is noticeably altered. While our learning-based approach cannot identify which specific component is most affected, the data shows that the system \emph{as a whole} is able to measure its temperature. And while a specialized temperature sensor would likely be more precise and targeted, we nonetheless believe that this proves the high versatility of the acoustic sensing approach, using only the embedded audio components to measure the approximate actuator temperature.

\subsection{Simultaneous Sensing of Location, Force, and Inflation With Classification Rates of \SI{95}{\percent}, \SI{97}{\percent}, and \SI{100}{\percent}}
\label{sub:combined_predictions}

The goal of this experiment is to show that contact location, contact force, and actuator inflation can be measured at the same time by the acoustic sensor. A single sound sample aggregates information about all three actuator parameters. By creating sensor models with different internal computations, we interpret the sensor signal in different ways to predict multiple actuator parameters simultaneously. At the same time, this experiment demonstrates that the sensor models are not affected by the other parameters. For example, the prediction of the contact location works regardless of the current contact force or actuator inflation.
This is important for applications in which multiple actuator properties need to be evaluated together to make a decision. For example, to evaluate the stability of a grasp, we may want to know the contact location as well as the contact force.

The data is recorded using the automated setup with the Panda robot and an active sound of \SI{20}{\milli\second} white noise. We use six contact locations (tip, middle, and base on both front and back of the finger) plus a seventh, ``no contact'' case. 
Each contact is recorded at two contact forces: \SI{1}{N} and \SI{3}{N}. The ``no contact'' case is considered as the third class with \SI{0}{N}. Furthermore, we record data at two inflation levels: \SI{0}{\kilo\pascal} and \SI{30}{\kilo\pascal}.  Each condition is recorded 25 times, for a total of 700 samples in the data set (7 locations $\times$ 2 forces $\times$ 2 inflation pressures $\times$ 25 repeats). 

For each property---location, force, and inflation---we create separate KNN sensor models that each use the complete training data. That means, for example, that each contact location class contains samples from all contact forces and all inflation pressures and vice versa. Each sensor model separately predicts the whole test set.

\begin{figure*}  
	\centering
	\begin{minipage}[t]{0.6\textwidth}
		\large{a} \hfilll \small{Contact Location} \hfilll \\
		{\hfilll \includegraphics[width=.85\textwidth,trim={0.5cm, 2cm, 0.5cm, 4.5cm},clip]{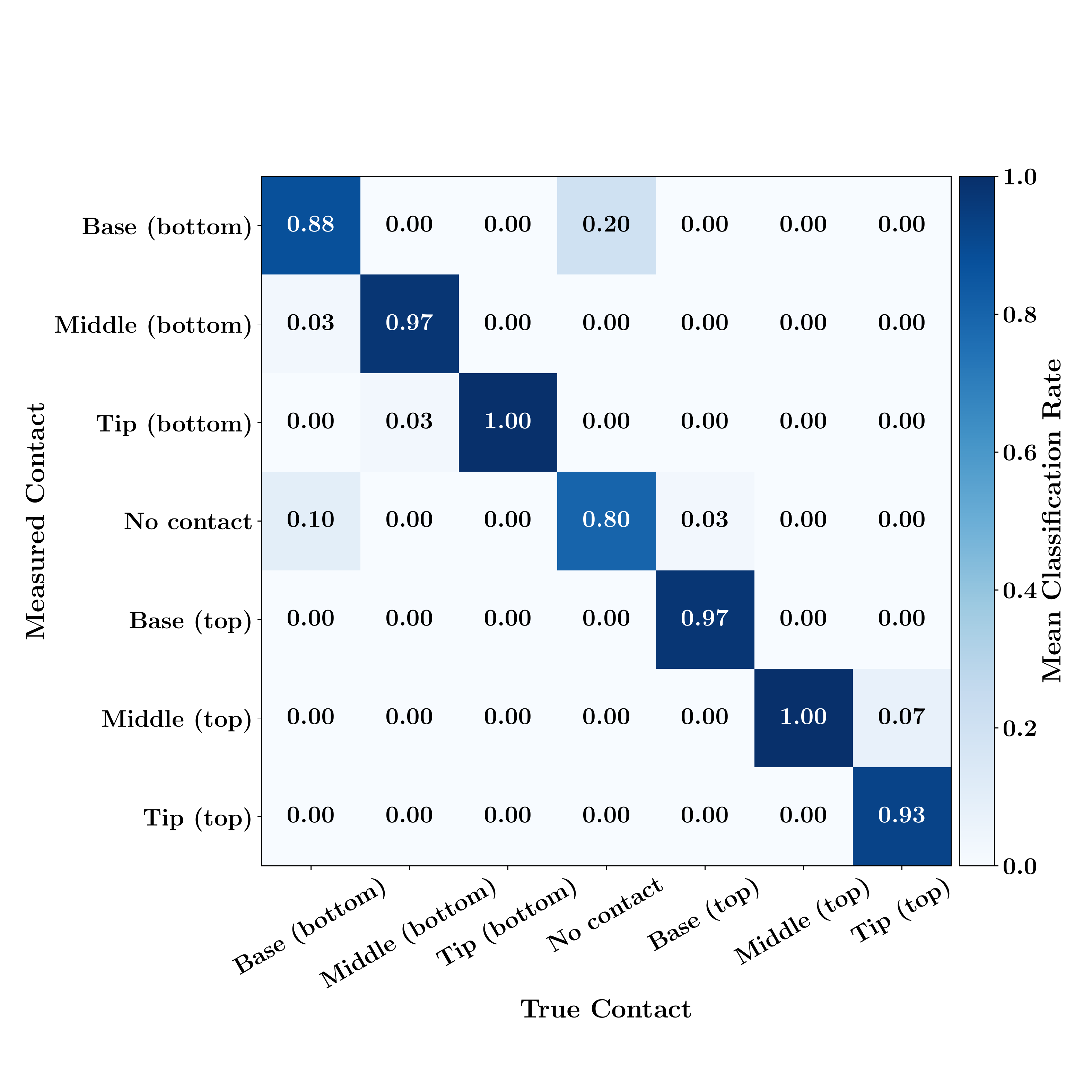} \hfilll}
	\end{minipage}
	\begin{minipage}[t]{0.35\textwidth}
		\large{b} \hfilll \small{Contact Force} \hfilll \\
		{\hfilll \includegraphics[width=0.8\textwidth, trim={0.5cm, 1.5cm, 0.5cm, 3cm}, clip]{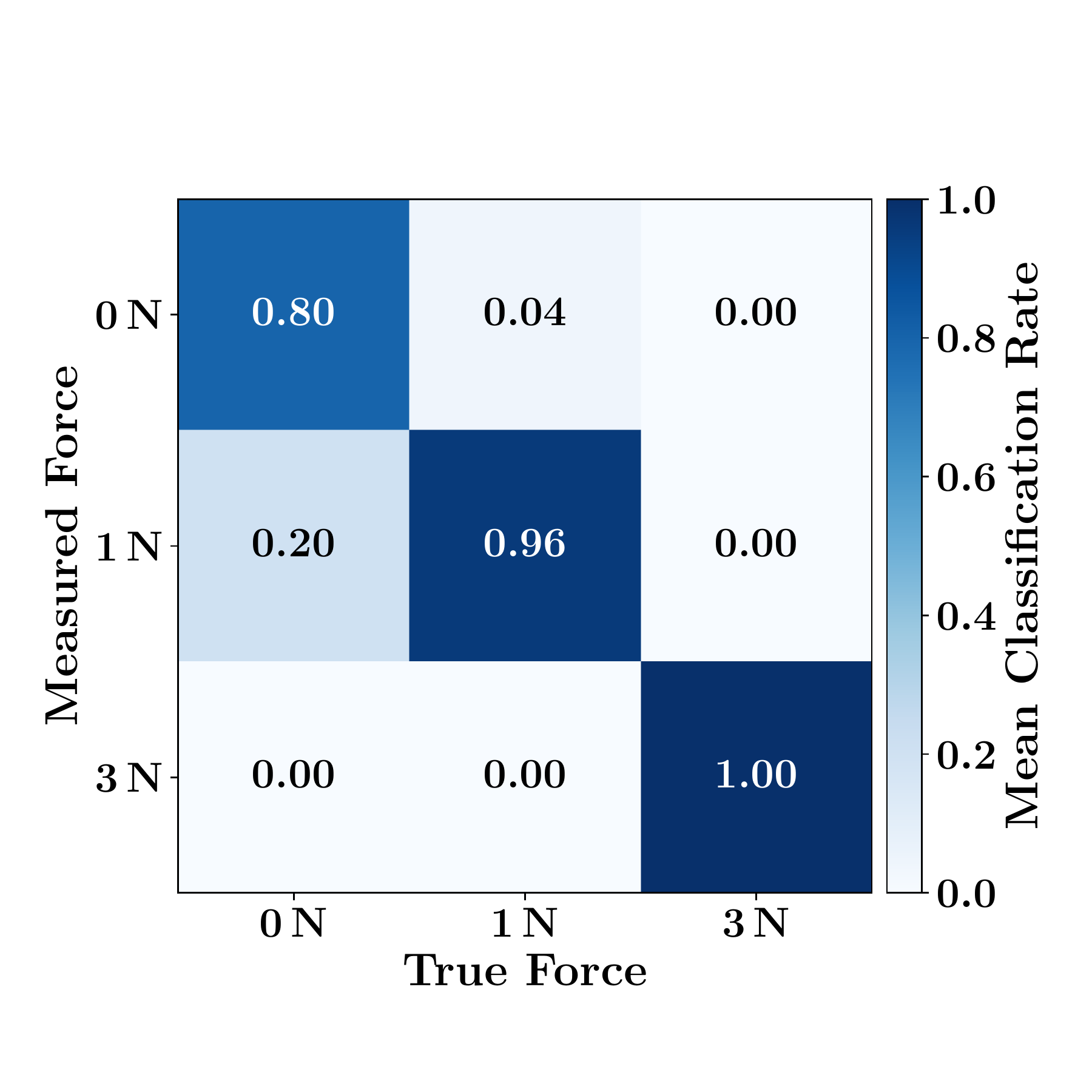} \hfilll \vspace{1em}}
		
		\large{c} \hfilll \small{Actuator Inflation} \hfilll \\
		{\hfilll \includegraphics[width=0.65\textwidth,trim={0.5cm, 2cm, 0.5cm, 2.5cm},clip]{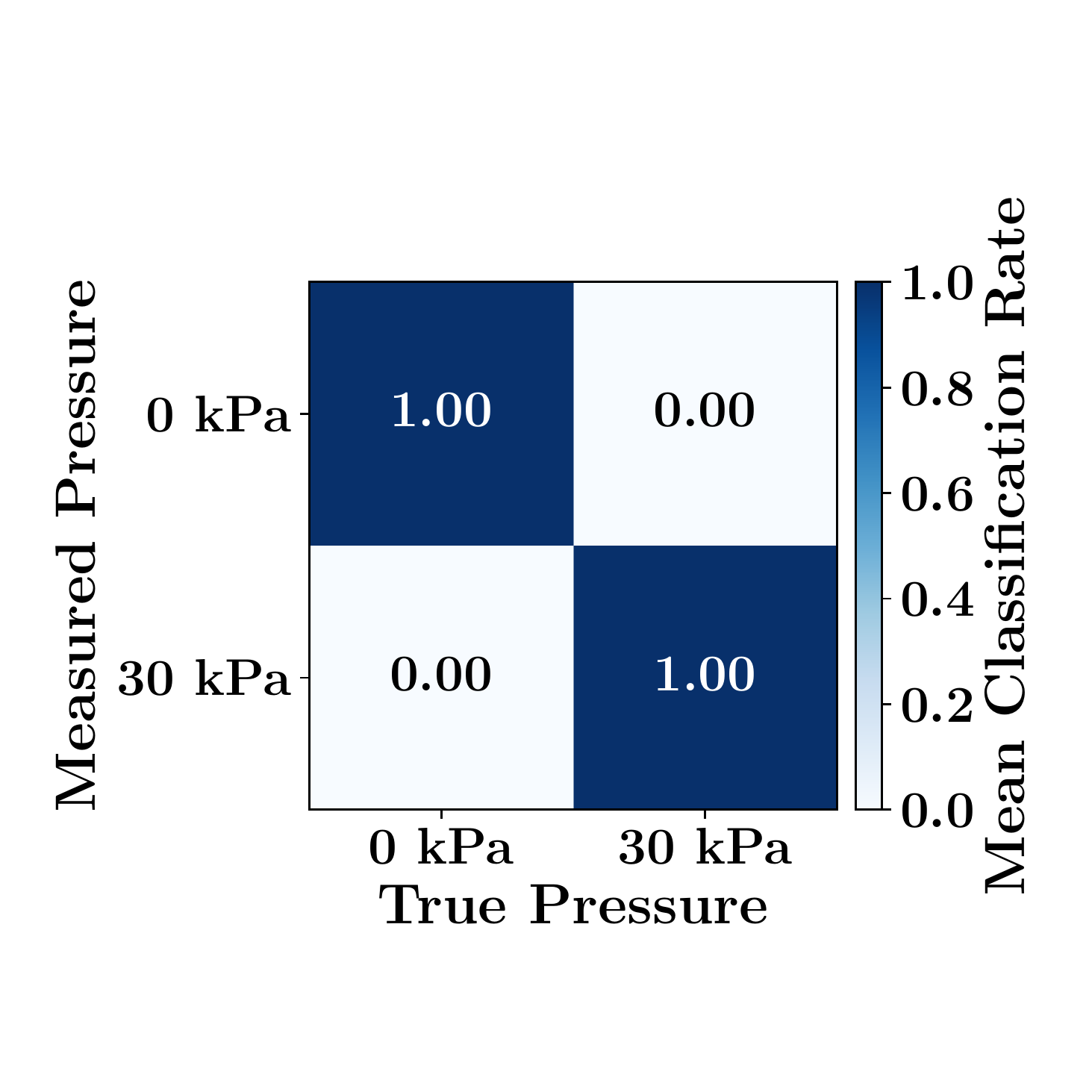} \hfilll}
	\end{minipage}

	\caption{Combined sensing of multiple properties: Using the same sound sample, we train separate sensor models to predict (a)~contact locations (ACR \SI{95}{\percent}), (b)~contact forces (ACR \SI{97}{\percent}), and (c)~inflation pressure (ACR \SI{100}{\percent}). These results demonstrate that different actuator properties can reliably be measured at the same time. 
	}
	\label{fig:exps_simultaneous}
\end{figure*}

Figures~\ref{fig:exps_simultaneous}a-c show the three confusion matrices for sensing the contact location, contact force, and inflation pressure. All plots show high values on the diagonals, indicating good prediction results. The contact location is predicted with a mean classification rate of \SI{95}{\percent}, which is slightly better than the active sensing results for contact locations in Section~\ref{sub:contact_location}, demonstrating that the active sensor is not negatively affected by having to generalize across different contact forces and inflation pressures. The same is true for the prediction of contact forces (\SI{97}{\percent} classification rate) and inflation pressures (\SI{100}{\percent} classification rate). This demonstrates that the same sound recording can be interpreted by different sensor models to measure three actuator properties simultaneously. And while this test cannot evaluate the \emph{influence} of each property on the sound, the fact that each model still achieves high classification rates regardless of the other properties shows that contact locations, contact forces, and actuator inflations each appear to modulate the sound along different, distinct frequencies, which do not interfere with the other sensor models.

\section{Acoustic Sensing is Robust Against Common Disturbances}
\label{sec:disturbances}

Successful sensorization for soft actuators must be robust to common disturbances. For acoustic sensing, external sounds or vibrations may be an issue. We show that acoustic sensing is robust to environmental noise, motor vibrations of the robot arm, as well as sounds by neighboring acoustic sensors. Given this robustness, acoustic sensing is well-suited for real-world robotic applications, even in noisy environments.

\subsection{No Influence of Background Noise}
\label{sub:background_noise}

To demonstrate the sensor's robustness against external noises, we repeat the location sensing experiments in the presence of loud noise and analyze the effect on the sensor's classification rate. If the background noise affects the sensor, we would expect to see a decrease in the classification rate as the noise levels increase.

The experimental setup is identical to the manual contact location recordings (Sec.~\ref{sub:contact_location}), with the addition of a pair of desktop speakers placed in a distance of \SI{10}{\centi\meter} from the sensorized actuator. The external speakers emit white noise at three volumes: \SI{50}{dB}, \SI{70}{dB}, and \SI{90}{dB}. 
We record a baseline data set in a quiet room (\SI{30}{dB}) and use it to train a location-predicting sensor model. With that model, we predict the ``noisy'' samples and report the average prediction accuracy across all contact locations. 
We repeat the experiment for the three sensing modes: passive, dynamic, and active.

Additionally, we estimate the signal-to-noise ratio (SNR) within the actuator by using the active recording as the \emph{signal} and the passive recording as the \emph{internal noise}. For the baseline in a quiet room, the SNR is \SI{47}{dB}. With increasing background noise, the SNR remains at \SI{47}{dB} and \SI{48}{dB}, before dropping slightly to \SI{43}{dB} for the loudest case. These values indicate that the influence of external noise on the internal recordings is minimal.

\begin{figure}
	\centering
	{\flushleft \large a\\}	
	\includegraphics[width=\linewidth]{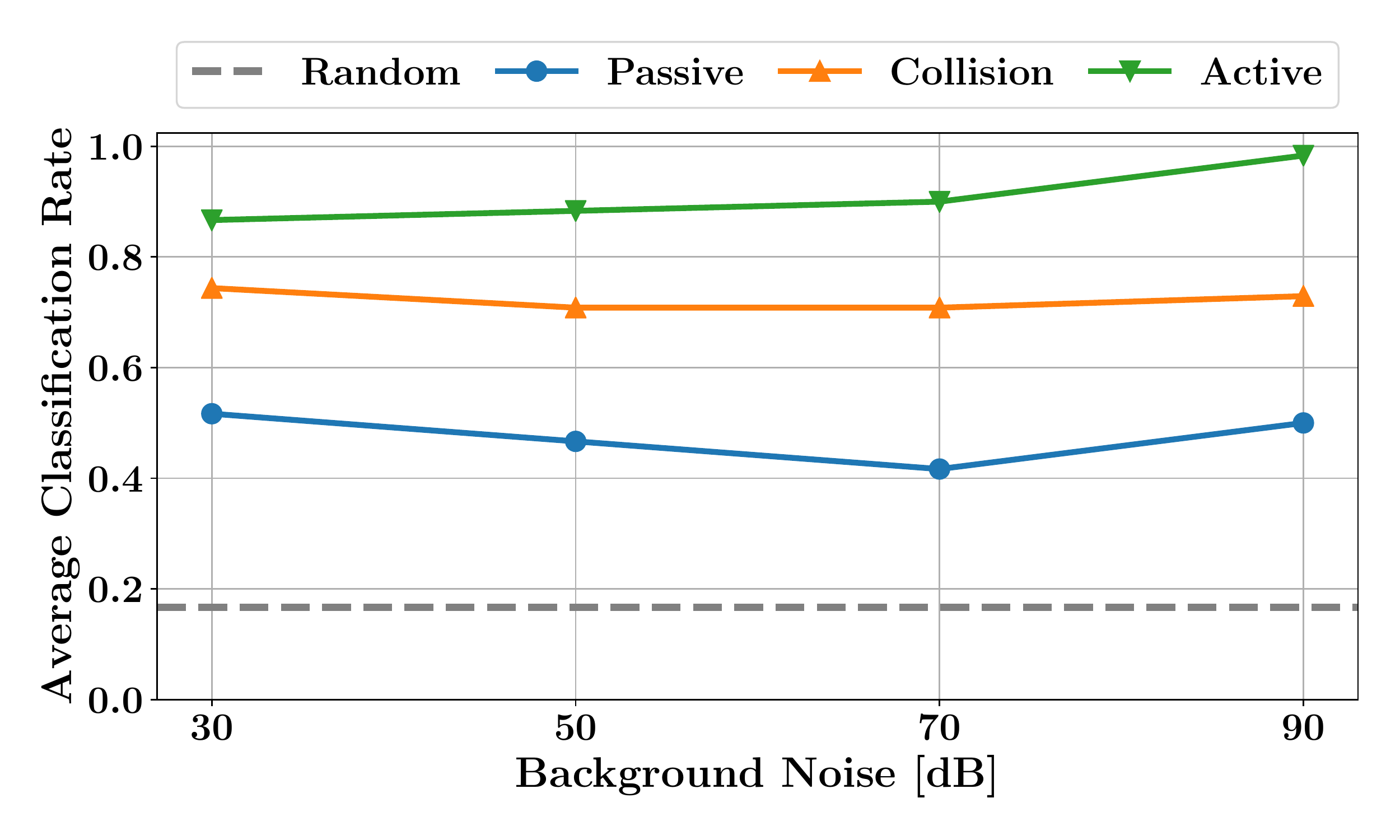}

	{\flushleft \large b\\}	
	\includegraphics[width=\linewidth]{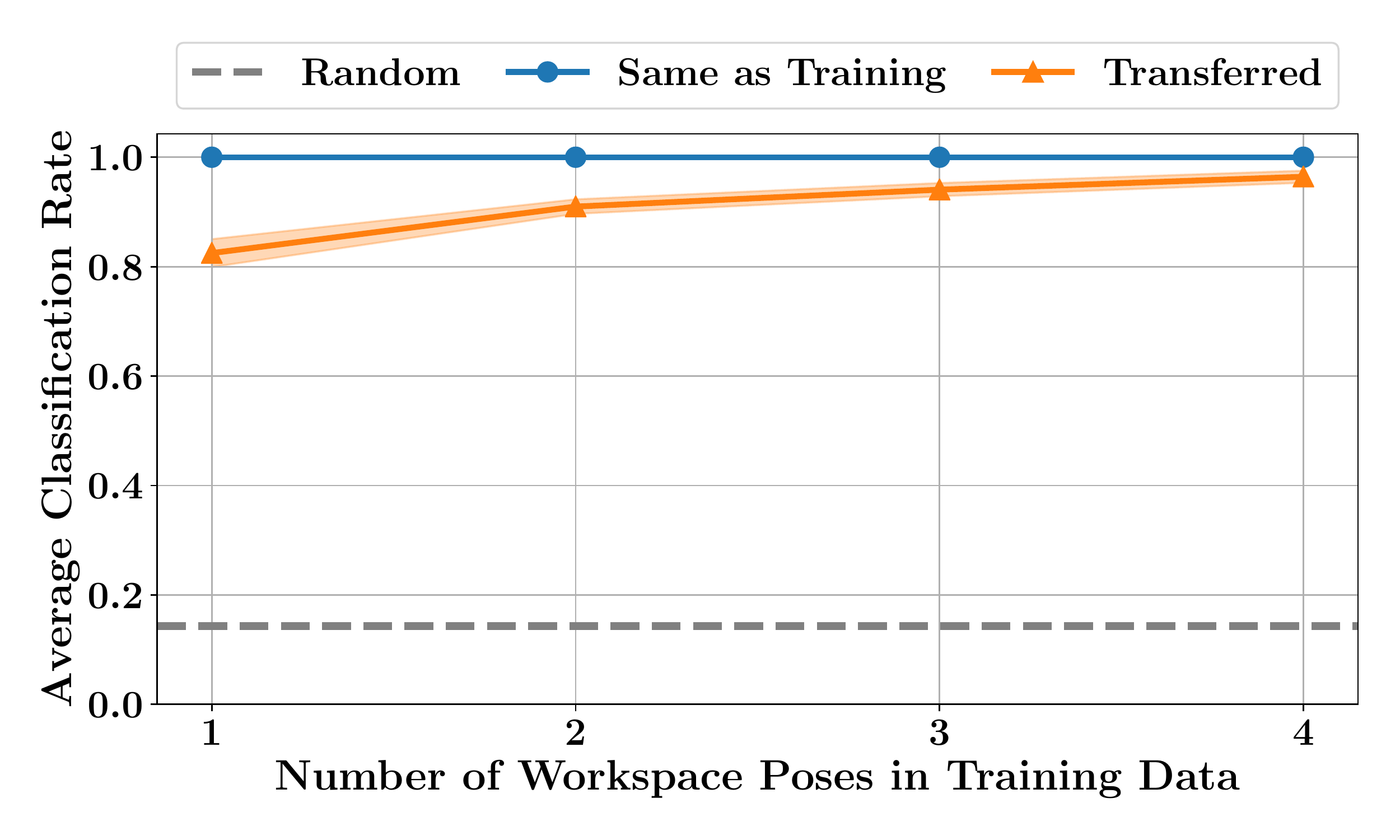}
	\caption{Robustness evaluation of the acoustic sensor: 
		\textbf{(a)}~Background noise does not affect acoustic sensing. Sensor models trained in a quiet room (\SI{30}{dB}) achieve identical or even slightly improved average prediction rates at high background noise levels (\SI{90}{dB}).
		\textbf{(b)}~Sensor models trained on a single workspace pose tend to overfit, i.e. high ACR on data from the same pose (blue line) and slightly lower ACR on other poses (orange line). By training on two or more different poses, the transfer to unseen poses is improved. The shaded area indicates the standard error of the mean across different pose combinations.}
	\label{fig:exp_noise_and_poses}
\end{figure}

This can also be seen in Figure~\ref{fig:exp_noise_and_poses}a, where the results show consistent classification rates
across all noise levels. This demonstrates that the acoustic sensor is unaffected by background noises. The active sensor's classification rate even increases slightly with louder noise, which might be due to the noise drowning out other, more systematic distractions. We believe that the silicone hull of the actuator acts as an acoustic insulator, shielding the embedded microphone and allowing the sensor to function, even for noise levels near the pain threshold of human hearing.

\subsection{Ignoring Pose-Specific Robot Noises}
\label{sub:vibrations}

The motors and gears of the robot arm, to which the sensorized actuator is mounted, produce noticeable vibrations. We show that the sensor model picks up on robot-pose specific noise patterns in the data, which leads to worse results when sensing in new poses. However, this can be improved by training on data from different robot poses.

We use the automated recording setup with a wooden contact object, an active sound of \SI{1}{\second} white noise, and seven contact classes (six contact locations, one ``no contact'' class). We record data for six workspace poses of the robot: the hand touches the object from the top, left, and right side, once close to the robot's base and once with the arm stretched out. For each workspace pose we record 175 samples. We train separate Support vector classifiers (SVC) using the parameters identified by the grid search in section~\ref{sub:ml_methods}. Additionally, we train classifiers on all combinations of data sets from two, three, and four different workspace poses. 

In Figure~\ref{fig:exp_noise_and_poses}b, we evaluate the sensor's robustness to pose variations by comparing the classification rates of test data from the \emph{same} workspace poses as the classifier's training data (blue, ``Same as Training'') and test data from the \emph{other} workspace poses, which the classifier has not seen before (orange, ``Transferred''). A pose-agnostic acoustic sensor should show little difference between the two cases.
It can be seen that for classifiers trained on a single workspace pose, the prediction of data from the same pose is at \SI{100}{\percent}, while the transfer to the other poses results in an average classification rate of only \SI{82}{\percent}. This indicates that the sensor model overfits to robot noises, i.e. the acoustic sensor learns to identify the pose-specific noises of the robot! However, when combining data from two or more different workspace poses, the transfer results improve up to \SI{96}{\percent} for four data sets. This shows that the acoustic sensor can learn to ignore pose-specific robot noises with more diverse training data.

\subsection{Neighboring Sensors Do Not Interfere}
\label{sub:interference}

A special case of extraneous noises are sounds from other active sensors. For example, a sensorized robot hand could have several fingers in close proximity, which may play the same active sounds. Nonetheless, we show that sounds from other active sensors do not interfere with the acoustic sensor.

We mount three additional sensorized fingers next to the sensorized index finger of the RBO Hand~2, each equipped with its own embedded speaker.
Using the automated recording setup with six contact locations on the index finger, we record 150 samples. All four fingers are active, i.e.\ all speakers simultaneously play the \SI{1}{\second} sweep sound. We train a KNN sensor model to classify the six contact locations on the index finger and compare the average classification rate to the location sensing results from the previous section.

Previously, when using only a single active finger, the mean contact location classification rate was up to \SI{100}{\percent} (see Sec.~\ref{sub:vibrations}). Now, when all four fingers are actively playing a sound, the acoustic sensor in the index finger still achieves a classification rate of \SI{96.7}{\percent}. This shows that neighboring active sensors do not significantly influence the classification rate, making it possible to use them in parallel on hands like the RBO Hand~2.

\section{Analyzing Sensor Design Choices}
\label{sec:method_parameters}

We now investigate how design choices for the computational acoustic sensor affect its performance. We analyze the influence of different types and volumes of active sounds on sensing performance, test the transferability of trained sensor models between PneuFlex actuators, and compare different machine learning methods for determining sensor models.

\subsection{Sensing Results Are Largely Unaffected by the Type and Duration of the Active Sound}
\label{sub:other_sounds}

An active acoustic sensor uses a sound to ``probe'' the actuator's state. We investigate how the choice of active sound affects sensing performance.

We compare four types of sounds:
\begin{enumerate}
  \item Logarithmic frequency sweep from \SI{20}{\hertz} to
    \SI{20}{\kilo\hertz}: This sound contains all frequencies of the speaker's
    range in sequential order. The logarithmic distribution emphasizes
    the lower frequencies, which we observed to be beneficial.
  \item White noise: This randomly generated sound has (statistically) uniform intensity across all
    frequencies. But unlike the sweep, there is no temporal order to the frequencies; they are shuffled randomly. For better comparability, the random signal is created once and then reused for all recordings.
  \item Band-limited white noise: This sound is based on white noise, but is
    bandpass-filtered to contain only frequencies between
    \SI{2}{\kilo\hertz} and \SI{4}{\kilo\hertz}. This frequency range
    corresponds to the biggest peak in the spectrum and contains
    noticeable shifts between contact classes. We suspect this to be
    the most relevant region.
  \item Sine wave with a frequency of \SI{2580}{\hertz}: To test if a
    single frequency might suffice, we generate a sine wave
    signal. Its frequency coincides with the biggest peak, i.e. resonance, in the
    recorded spectra.
\end{enumerate}
    
In addition to the four sound types, we also investigate the effect of the sound duration. Each sound is evaluated in five lengths: \SI{5}{\milli\second}, \SI{20}{\milli\second}, \SI{50}{\milli\second}, \SI{500}{\milli\second}, and \SI{1}{\second}. 
For each sound type and duration, we repeat the automated contact location experiment, with seven contact classes recorded using the Panda robot arm. Each data set consists of 175 samples. We use the average test set classification rate to judge the success of each sound. 

\begin{figure}
	\centering
	{\flushleft \large a\\}	
	\includegraphics[width=\linewidth, trim={0.5cm 0.7cm 0.5cm 0.7cm}, clip]{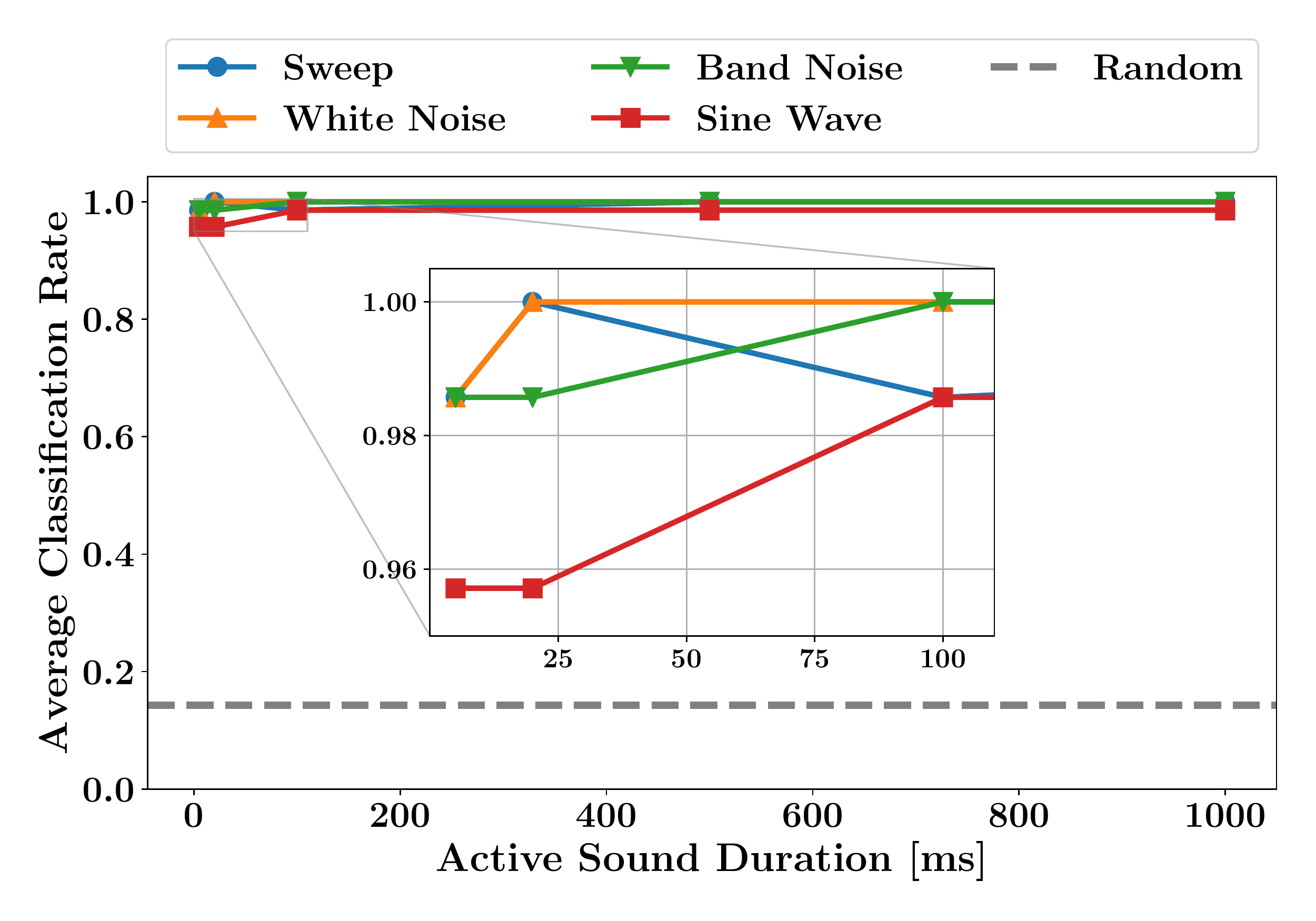}
	{\flushleft \large b\\}	
	\includegraphics[width=\linewidth, trim={0.5cm 0.7cm 0.5cm 0.7cm}, clip]{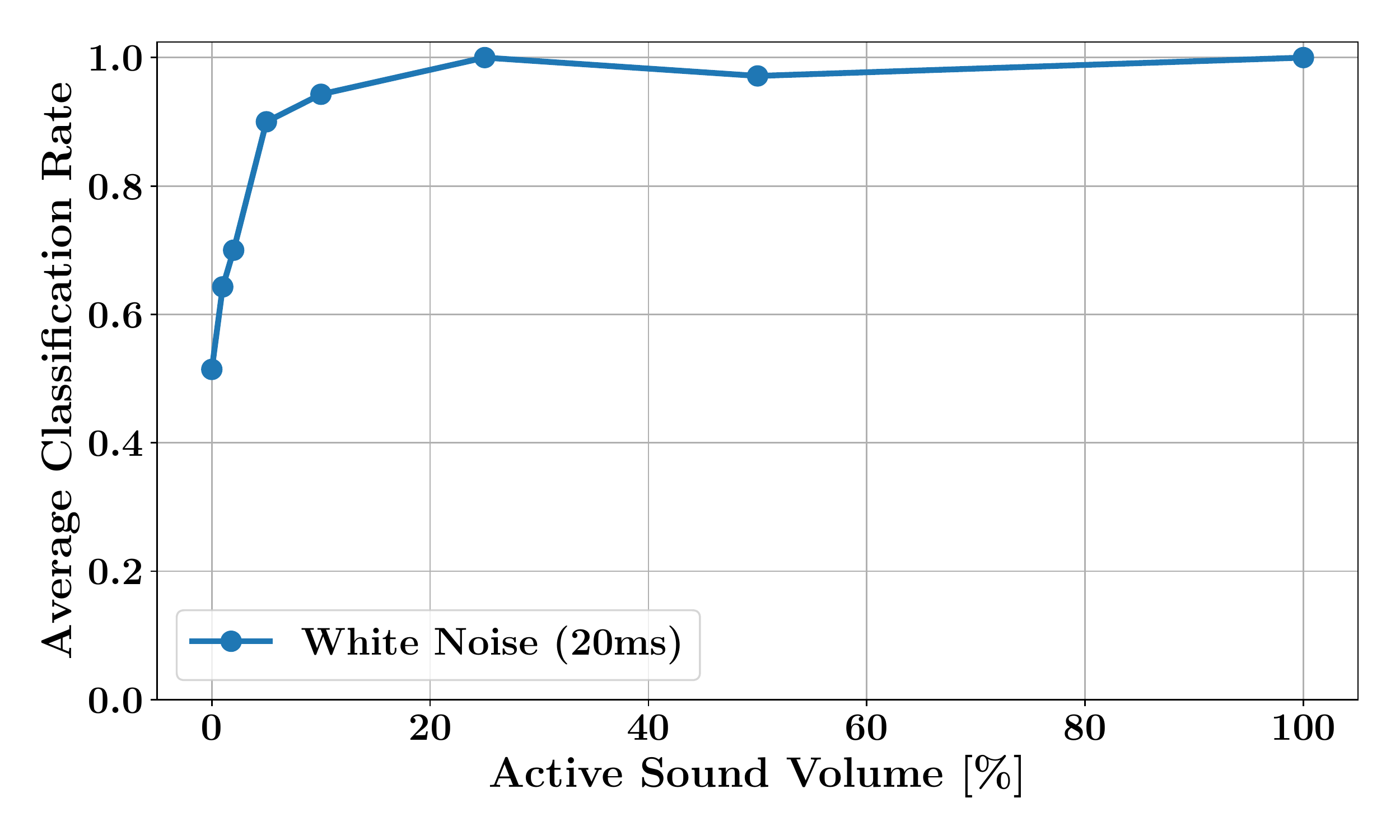}
	\caption{Evaluation of different active sounds: 
		\textbf{(a)}~The sensing performance is largely independent of the specific type and duration of the sound played by the embedded speaker. 
		\textbf{(b)} By reducing the volume of the active sound to \SI{25}{\percent}, we minimize the external noise while maintaining high sensing accuracy.}
	\label{fig:experiments_3a}
\end{figure}

The plot in Figure~\ref{fig:experiments_3a}a shows the average classification rates for the four sound types and five durations. There is little difference between the different types of sound, with the single sine wave achieving the lowest average of \SI{97.4}{\percent} and white noise the highest average of \SI{99.7}{\percent}. Similarly, the duration of the sound has almost no influence, with only a small decrease of the classification rate for the shortest sounds at \SI{5}{\milli\second}.
Overall, these experiments show that the active acoustic sensor's ability to identify the actuator's state-dependent modulation is largely independent of the type and duration of the sound played. However, the best results are achieved by wide-frequencies sounds, like sweep and white noise, and a duration of at least \SI{20}{\milli\second}.

\subsection{Sensing Performance Remains High at \SI{25}{\percent} Sound Volume}
\label{sub:volume}

We initially set the sound volume of the embedded speaker to the highest value that did not create clipping in the microphone. This maximizes the detail in the recorded samples, making it easier to identify the state-dependent changes in sound modulation. At that level, the active sound is audible on the outside of the actuator. We investigate how the sound volume affects the prediction accuracy, to find a value that minimizes noise while keeping the sensor performance high.

We define the maximum volume (\SI{100}{\percent}) as the loudest speaker signal without any clipping in the microphone signal. Additionally, we record data sets at \SI{50}{\percent}, \SI{25}{\percent}, \SI{10}{\percent}, \SI{5}{\percent}, \SI{2}{\percent}, \SI{1}{\percent}, and \SI{0}{\percent}. The latter is identical with \emph{passive} sensing, where no active sound is emitted. 
The sound type is white noise with a duration of \SI{20}{\milli\second}. The automated setup is used to record 25 samples at six contact locations, for a total of 150 samples at each sound volume.
Separate sensor models are trained for each volume and compared via the average contact location classification rate. 

Figure~\ref{fig:experiments_3a}b shows that the prediction rate remains high for sound volumes of \SI{25}{\percent} and higher. Below that it starts to drop rapidly. This logarithmic profile corresponds nicely to the decibel scale of the sound pressure level and indicates that the sensor's performance depends on the energy of the active signal. 
At \SI{25}{\percent} of the maximum sound volume, the acoustic sensor still maintains its high sensing accuracy, while minimizing its external noise.

\subsection{Actuator-Specific Acoustic Properties Make Transfer of Sensor Models Difficult}
\label{sub:model_transfer}

The acoustic properties of different actuators will likely differ to some degree as a result of the manufacturing process. To evaluate how this affects the transferability of trained sensor models between actuators, we record data for five actuators and compare their sensor models.

Using the automated experimental setup, we record samples for four contact classes (tip, middle, base, and no contact) for the five actuators. Each class is sampled 25 times, for 100 samples in total per actuator. The active sound is \SI{20}{\milli\second} of white noise.
For each actuator, we train a separate sensor model using 60 training samples and the default KNN classifier. Additionally, we create four combined models, trained on data from two or three different actuators.

\begin{figure}
	\centering
	{\flushleft \large a\\}	
	\includegraphics[width=\linewidth]{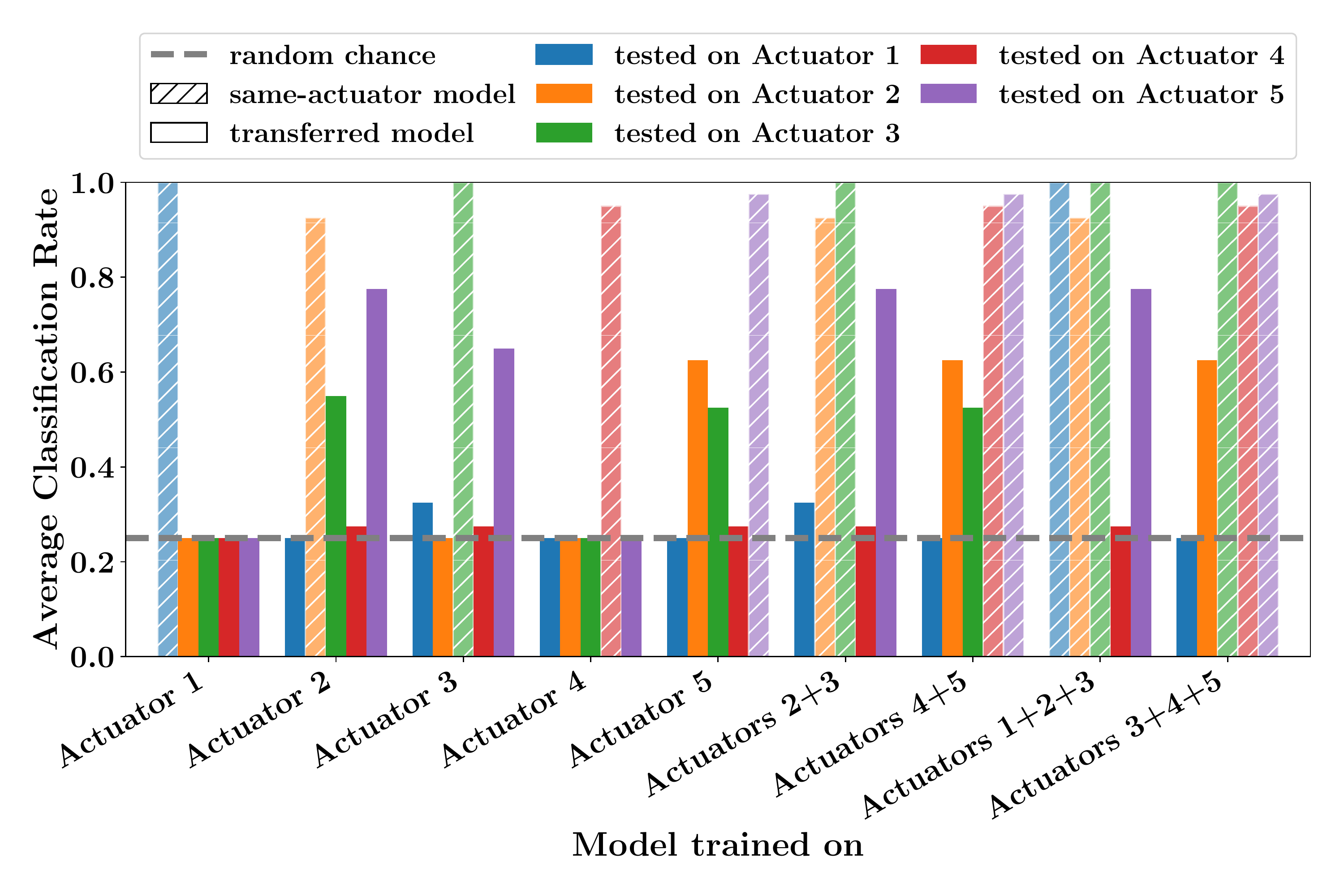}
	
	{\flushleft \large b\\}	
	\includegraphics[width=0.75\linewidth]{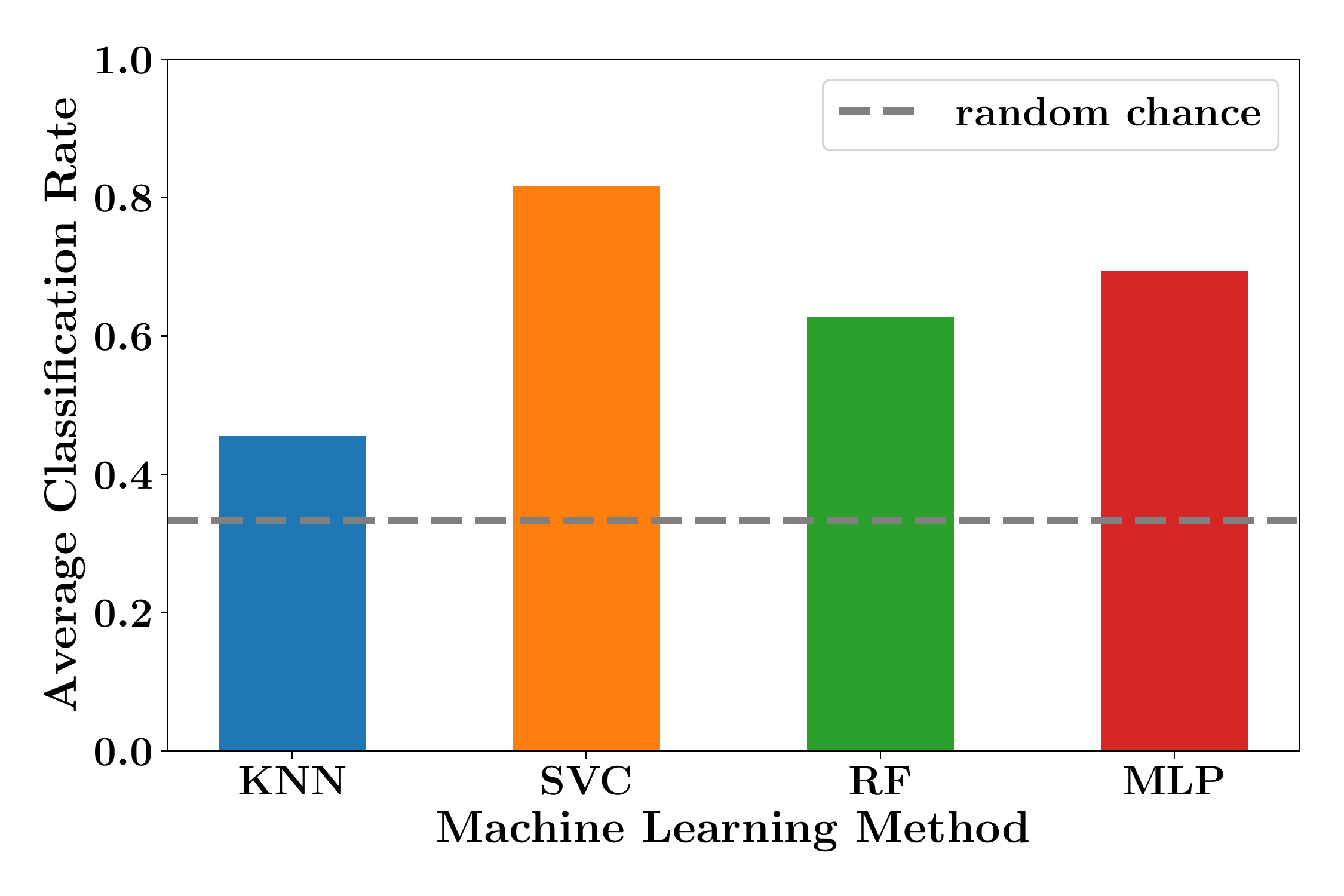}

	\caption{Evaluation of model transfer and different learning methods: 
		\textbf{(a)}~Sensor models do not transfer well between different actuators. Cross-actuator sensing (solid bars) achieves classification rates close to random guessing. When training sensor models on data from multiple actuators we achieve a small improvement for cross-actuator prediction.
		\textbf{(b)}~Grid search result for material sensing: While the KNN model is only slightly better than random chance, other machine learning methods achieve high classification rates. This indicates that more complex computation helps to extract the relevant information from the modulated sound.}
	\label{fig:experiments_3b}
\end{figure}

The results in Figure~\ref{fig:experiments_3b}a show high classification rates for same-actuator measurements, i.e. for the actuators the models were trained on, with an average of \SI{97}{\percent} for both single-actuator and multi-actuator models. However, the cross-actuator measurements, i.e. for data from actuators the models were not trained on, are significantly worse. At \SI{35}{\percent} ACR the single-actuator transfer result is just above the random guessing baseline of \SI{25}{\percent}. When training on more than one actuator, the model transfer results are improved slightly, but with \SI{47}{\percent} ACR still comparatively low.
This indicates that the acoustic differences between actuators are significant and sensor models trained on one actuator do not transfer well to others. In the future, we plan to identify a calibration procedure to normalize the spectra of each actuator. For now, we create separate sensor models for each actuator, similar to the factory-calibration of other sensors.

\subsection{Improving Sensing Results With More Complex Learning Methods}
\label{sub:ml_methods}

One of the key components of the acoustic sensor is the sensor model's mapping from sound to measurement. In most experiments, we used a basic k-nearest neighbor classifier and achieved high classification rates for a large range of measurable properties.
To see if more complex predictors will further improve the sensor measurements, we now compare different machine learning methods. As a benchmark problem, we use the sensing of ``object materials'' (Sec.~\ref{sub:material}) for which the basic KNN classifier did not perform well. For all other problems, the classification rates were high and we would expect less distinctive results.

We reuse the data recorded for the ``sensing object material''-experiment and perform individual grid searches for the following learning methods (searched parameters in brackets, best in \textbf{bold}):
\begin{enumerate}
	\item K-nearest neighbor classifier (KNN)\\ {\small n\_neighbors=[1, 2, \textbf{3}, 5, 10], distance\_metric=[L1, \textbf{L2}]}
	\item Support vector classifier (SVC)\\ {\small (kernel=[\textbf{linear}, rbf], C=$10^{[\textbf{-2}..10]}$, gamma: [\textbf{scale}, auto])}
	\item Random forest (RF)\\ {\small (n\_estimators=[10, 50, 100, \textbf{500}], max\_features=[\textbf{sqrt}, log2], \\
	max\_depth=[5, 10, \textbf{full}])}
	\item Multi-layer perceptron (MLP) \\ {\small hidden\_layers=[\textbf{(100)}, (200,200), (300,300,300)], alpha=[0.001, \textbf{0.1}, 1]}
\end{enumerate}
For each method, we identify the best parameters using scikit-learn's cross-validated grid search function on the 270 training samples. Each best model is evaluated on a separate test set of 180 samples.

The ACR of each method's best parameter model is shown in Figure~\ref{fig:experiments_3b}b. 
The material sensing results of the basic KNN model are significantly improved by the other models. The support vector classifier achieves the highest score with an ACR of \SI{82}{\percent}. This indicates that we can further improve the performance of the acoustic sensor by using more complex computation methods. Hence, the already very good results of the KNN models can be considered as lower bounds of what the acoustic sensor can measure. 



\section{Limitations of Acoustic Sensing}
\label{sec:limitations}

In our experiments, acoustic sensing has proven to be a simple, versatile, and robust approach to the sensorization of soft actuators. We now discuss when it \emph{should not} be used and which aspects still require further research. 

Because the sensorization method is based on recognizing small variations in the recorded sound, it may fail when objects produce sound themselves. Upon contact with such an object, the sound could be transmitted into the actuator's air chamber and get recorded by the microphone. If that sound is not in the training data, the sensor model will likely be unable to extract the correct actuator measurement.

Another limitation results from the poor transferability of sensor models across different actuators. This could possibly be overcome with a short routine that consists of playing specific ``calibration sounds'' to identify and compensate acoustic differences. This is similar to the initial factory calibration of other sensors. However, this makes acoustic sensing a less desirable technology for applications in which sensorized parts of the robot have to be replaced routinely.

Our acoustic sensor relies on the observable modulation of sound. But not all actuator states and interactions affect the modulation equally. While some actuator properties, e.g.~contact location, create distinct sound modulations which enable high classification rates, other properties, like the object material, appear to affect the sound modulation less and may require more complex machine learning techniques to detect reliably. It is important to identify a sensor model that works well for a desired measurement.

Novel state properties might become accessible with different representations of the recorded sounds. We currently use simple frequency spectra, which are fast to compute and appear to contain relevant information. Other acoustic sensing approaches successfully employed different representations. Spectrograms, for example, additionally contain information about frequency attenuation~\citep{krotkov_robotic_1997,harrison_tapsense_2011,mikogai_contact_2020}. 
The exploration of different sound representations and matching learning methods may lead to additional actuator properties becoming measurable.

Finally, there is the number of simultaneously measurable actuator parameters. We showed that a single acoustic sensor can reliably measure the contact location, contact force, and actuator inflation at the same time (Sec.~\ref{sub:combined_predictions}). It should be possible to include additional measurements, as long as their sound modulations are unambiguous. However, the current implementation requires recording training samples for each parameter combination. To avoid this combinatorial explosion, it might be beneficial to use a hierarchical sensor structure instead. The output of one sensor model could be used as a simplifying prior to the next model. Future research will have to determine the best structure of such a hierarchical sensor network.

\section{Conclusion}

We proposed active and passive acoustic sensing as a simple, robust, and versatile sensorization method for soft actuators. As sound travels through the actuator, it is modulated depending on the actuator's current physical state (e.g. shape, forces, contact). From small changes in the sound's frequency spectrum, we infer the corresponding actuator property using machine learning. Our acoustic sensor thus consists of physical components which record the modulated sound (embedded microphone and speaker), and a computational component which extracts the desired measurement from sound (trained sensor model). Such a ``computational sensor'' can use the same physical hardware to emulate a range of special-purpose sensors, like contact and force sensors. This acoustic sensing principle has a wide range of possible applications, especially in soft robotics where sensors need to be flexible and functional.

We demonstrated the effectiveness of acoustic sensing in the context of a soft, pneumatic PneuFlex actuator. The sensor achieved reliable and accurate measurements for contact location, contact force, and actuator inflation. From a single sound recording, all three properties can be predicted simultaneously. Additionally, the sensor was capable of recognizing the material of contact objects and measuring the actuator's temperature, all from recordings of the internal sound. At the same time, the rubber hull of the actuator shields the microphone from external sounds, so that even loud background noises do not affect the measurements. All this makes acoustic sensing a versatile approach for the sensorization of soft pneumatic actuators. 
 
\begin{acks}
The authors wish to thank Marius Hebecker for helping with the robot experiments.
\end{acks}

\begin{funding}
This research was funded by the European Commission (SOMA, H2020-ICT-645599), the Deutsche Forschungsgemeinschaft (DFG, German Research Foundation) under Germany's Excellence Strategy - EXC 2002/1 "Science of Intelligence" - project number 390523135 and German Priority Program DFG-SPP 2100 ``Soft Material Robotic Systems''.
\end{funding}

\begin{dci}
The authors declare that there is no conflict of interest.
\end{dci}


\flushend

\bibliographystyle{SageH}


\end{document}